\documentclass[letterpaper, 10 pt, conference]{ieeeconf}  
\IEEEoverridecommandlockouts                              
\usepackage{graphicx} 
\usepackage{todonotes}
\usepackage{pgfplots}
\pgfplotsset{compat=newest}
\usepackage{subfig}
\usepackage{multirow}
\usepackage{tabularx}
\usepackage{hyperref}
\usepackage{array}
\usepackage{float}
\usepackage{graphbox}
\usepackage{tikz}
\usetikzlibrary{positioning,shapes}
\usetikzlibrary{calc,trees,positioning,arrows,chains,shapes.geometric,%
	decorations.pathreplacing,decorations.pathmorphing,shapes,%
	matrix,shapes.symbols,automata}
\usepackage{flushend}
\usepackage[binary-units=true]{siunitx}
\usepackage{amsmath} 
\usepackage{amssymb}  

\newcommand{\norm}[1]{\left\lVert#1\right\rVert}
\newcommand{\abs}[1]{\lvert#1\rvert}
\DeclareMathOperator*{\argmin}{arg\,min}

\newcommand{\reffig}[1]{Fig.~\ref{#1}}
\newcommand{\reftab}[1]{Tab.~\ref{#1}}

\newcommand{\etal}{et al.~}

\usepackage[font=small]{caption}

\definecolor{dark_green}{RGB}{43 200 0}
\makeatletter
\pgfplotsset{
	/tikz/max node/.style={
		anchor=south,
	},
	/tikz/min node/.style={
		anchor=north,
		name=minimum
	},
	mark min/.style={
		point meta rel=per plot,
		visualization depends on={x \as \xvalue},
		scatter/@pre marker code/.code={%
			\ifx\pgfplotspointmeta\pgfplots@metamin
			\def\markopts{}%
			\coordinate (minimum);
			\node [min node] {
			};
			\else
			\def\markopts{mark=none}
			\fi
			\expandafter\scope\expandafter[\markopts,every node near coord/.style=green]
		},%
		scatter/@post marker code/.code={%
			\endscope
		},
		scatter,
	},
	mark minGreen/.style={
		point meta rel=per plot,
		visualization depends on={x \as \xvalue},
		scatter/@pre marker code/.code={%
			\ifx\pgfplotspointmeta\pgfplots@metamin
			\def\markopts{}%
			\coordinate (minimum);
			\node [min node] {
			};
			\else
			\def\markopts{mark=none, color=black}
			\fi
			\expandafter\scope\expandafter[\markopts,every node near coord/.style=green, color=dark_green, mark size=3pt]
		},%
		scatter/@post marker code/.code={%
			\endscope
		},
		scatter,
	},
	mark max/.style={
		point meta rel=per plot,
		visualization depends on={x \as \xvalue},
		scatter/@pre marker code/.code={%
			\ifx\pgfplotspointmeta\pgfplots@metamax
			\def\markopts{}%
			\coordinate (maximum);
			\node [max node] {
				(\pgfmathprintnumber[fixed]{\xvalue},%
				\pgfmathprintnumber[fixed]{\pgfplotspointmeta})
			};
			\else
			\def\markopts{mark=none}
			\fi
			\expandafter\scope\expandafter[\markopts]
		},%
		scatter/@post marker code/.code={%
			\endscope
		},
		scatter
	}
}
\makeatother

\title{\LARGE \bf Beyond Photometric Consistency: Gradient-based Dissimilarity \\for Improving Visual Odometry and Stereo Matching}

\author{Jan Quenzel \and Radu Alexandru Rosu \and Thomas L{\"a}be \and Cyrill Stachniss \and Sven Behnke
\thanks{This work has been supported as part of the research group FOR 1505
by the Deutsche Forschungsgemeinschaft (DFG, German Research Foundation) as well as under Germany's Excellence Strategy, EXC-2070 - 390732324 (PhenoRob).
Jan Quenzel, Radu Alexandru Rosu and Sven Behnke are with the Autonomous Intelligent Systems Group, University of Bonn, Germany.
Thomas L{\"a}be and Cyrill Stachniss are with the Robotics and Photogrammetry Lab,
Institute of Geodesy and Geoinformation, University of Bonn, Germany.
		}%
}

\begin{document}

\maketitle

\thispagestyle{empty}
\pagestyle{empty}

\begin{tikzpicture}[remember picture,overlay]
\node[anchor=north west,align=left,font=\sffamily,yshift=-0.2cm] at (current 
page.north west) {%
  In: International Conference on Robotics and Automation (ICRA), 2020
};
\node[anchor=north east, align=right,font=\sffamily,yshift=-0.2cm] at (current 
page.north east) {
};
\end{tikzpicture}%

\begin{abstract}
Pose estimation and map building are central ingredients of autonomous robots
and typically rely on the registration of sensor data. In this paper, we investigate
a new metric for registering images that builds upon on the idea of the photometric error.
Our approach combines a gradient orientation-based metric with a magnitude-dependent scaling term.
We integrate both into stereo estimation as well as visual odometry systems and show clear 
benefits for typical disparity and direct image registration tasks when using our proposed metric. 
Our experimental evaluation indicats that our metric leads to more robust
and more accurate estimates of the scene depth as well as camera trajectory. Thus, the metric improves camera pose estimation and in turn the mapping capabilities of mobile robots. We believe that a series of
existing visual odometry and visual SLAM systems can benefit from the findings reported in this paper.
\end{abstract}

\section{Introduction}

The ability to estimate the motion of a mobile platform based on onboard sensors is a key
capability for mobile robots, autonomous cars, and other intelligent vehicles. Computing
the trajectory of a camera is often referred to as visual odometry or VO and several approaches
have been presented in this context~\cite{LSDSLAM,kerl13icra,SVO,ORBSLAM2,DSO}.  VO as well as 
stereo matching approaches should provide accurate estimates of the relative camera 
motion and scenes depth under various circumstances. Thus, optimizing such systems
towards increased robustness is an important objective for robots operating in the real world.

The gold standard for computing the relative orientation of two images of a calibrated camera is  Nister's 5-point algorithm~\cite{nister2004fivePoint}. This approach computes the 5-DoF transformation between two monocular images based on known feature correspondences. It requires at least five corresponding points per image pair. In practice, more points are required to combine the 5-point algorithm with RANSAC followed by a least-squares refinement using only the inliers correspondences. An alternative approach to using explicit feature correspondences are comparisons of the pixel intensity values within the image pair. This approach is also called direct alignment and one often distinguishes semi-dense and dense methods, depending on the amount of compared pixels~\cite{DSO,SVO2,BCAVariantenPaper}.

Features are often designed to be resilient against changes in the intensity values of the images, for example caused by illumination changes. Often, features are  sparsely distributed over the image and their extraction can be a time consuming operation. In contrast to that, the intensity values of each pixel are directly accessible, raw measurements, and can be compared easily. Several direct methods consider the so-called photometric consistency of the image as the objective function to optimize. A key challenge of direct approaches is to achieve robustness because slight variations of the camera exposure, illumination change, vignetting effects, or motion blur directly affect the intensity measurements. In this paper, we address the problem of robustifying the direct alignment of image pairs through a new dis-similarity metric and in this way enable an improved depth estimate and alignment of image sequences.

The main contribution of this paper is a novel metric for direct image alignment and its exploitation in direct visual odometry. We build upon the gradient orientation-based metric proposed by Haber and Modersitzki~\cite{NGF} and improve it through the introduction of a magnitude depending scaling term. We furthermore integrate our metric into four different estimation systems (OpenCV, MeshStereo, DSO and Basalt) to show that our metric leads to improvements and evaluate our system to support our key claims, which are: 
First, our proposed metric is better suited for stereo disparity estimation than existing approaches.
Second, it is also well-suited for direct image alignment.
Third, our metric can be integrated into existing VO systems and increase their robustness while running at the frame rate of a typical camera.

\bgroup
\newcommand{\qualitativeElemWidth}{2.15cm}
\renewcommand{\tabcolsep}{1pt}
\begin{figure*}
	\centering
	\begin{tabular}{ccccccccc}
        RGB image&Modified RGB&$e_\mathit{photo}$& $e_\mathit{ncc}$ & $e_\mathit{mag}$ & $e_\mathit{gom}$ & $e_\mathit{ugf}$ &$e_\mathit{sgf}$ \\
		\includegraphics[width=\qualitativeElemWidth{}]{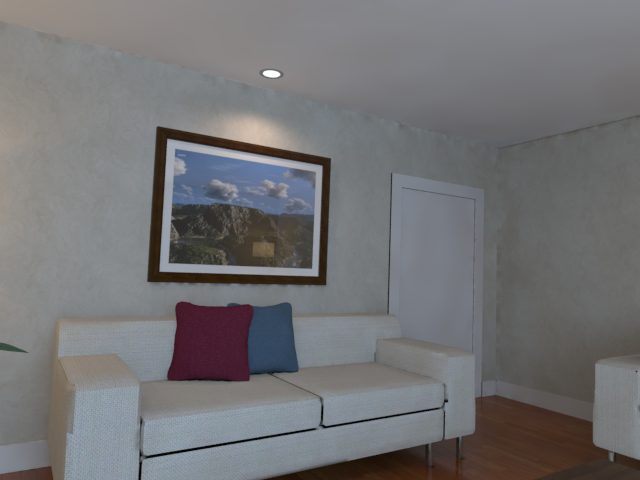}& 
		\includegraphics[width=\qualitativeElemWidth{}]{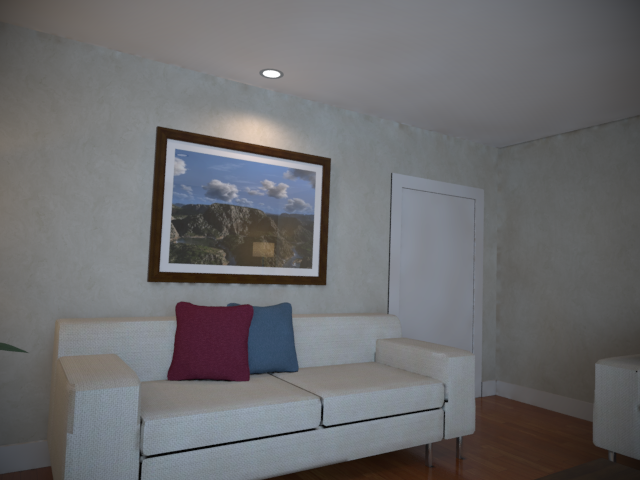}&
		\includegraphics[width=\qualitativeElemWidth{}]{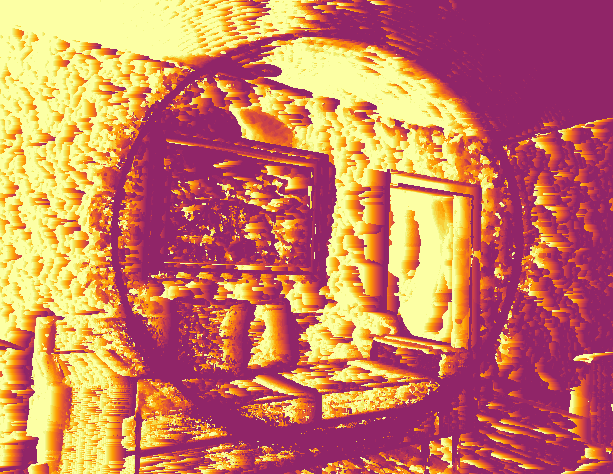}&
		\includegraphics[width=\qualitativeElemWidth{}]{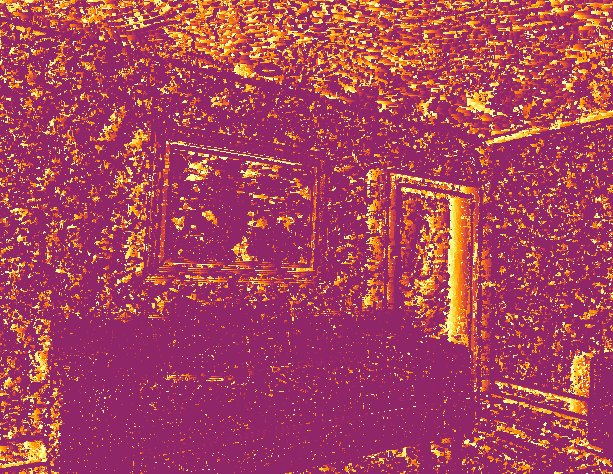}&
		\includegraphics[width=\qualitativeElemWidth{}]{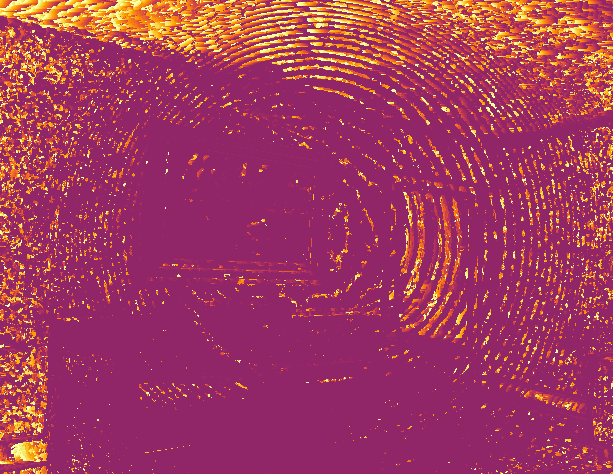}&
		\includegraphics[width=\qualitativeElemWidth{}]{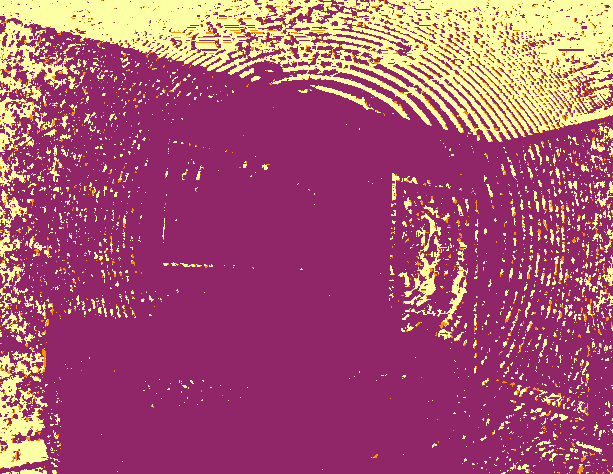}&
		\includegraphics[width=\qualitativeElemWidth{}]{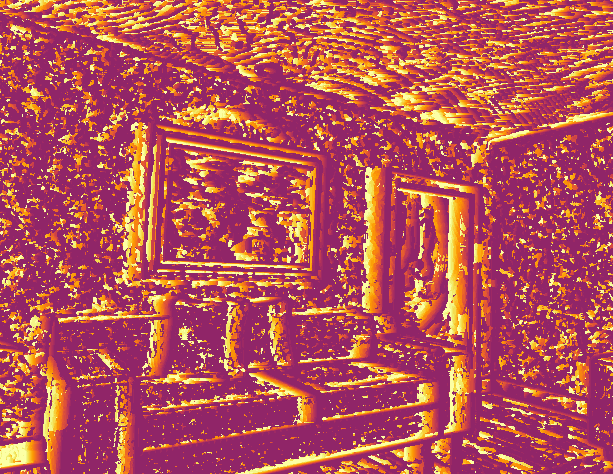}&
		\includegraphics[width=\qualitativeElemWidth{}]{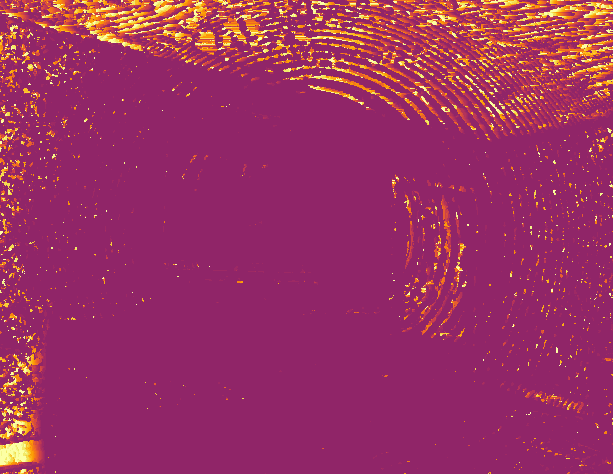}\\	
	
		\includegraphics[width=\qualitativeElemWidth{}]{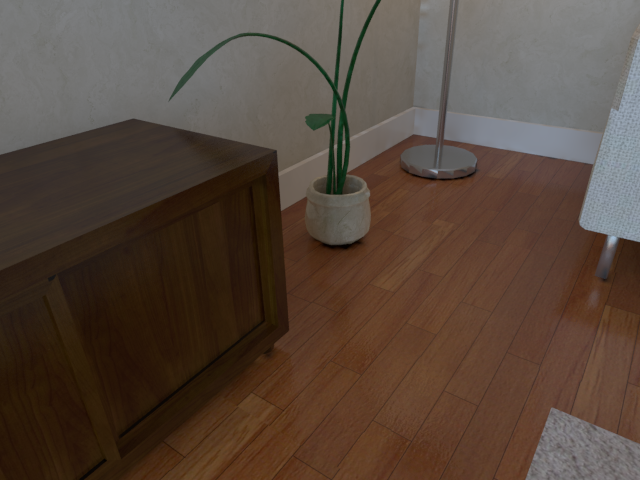}& 
		\includegraphics[width=\qualitativeElemWidth{}]{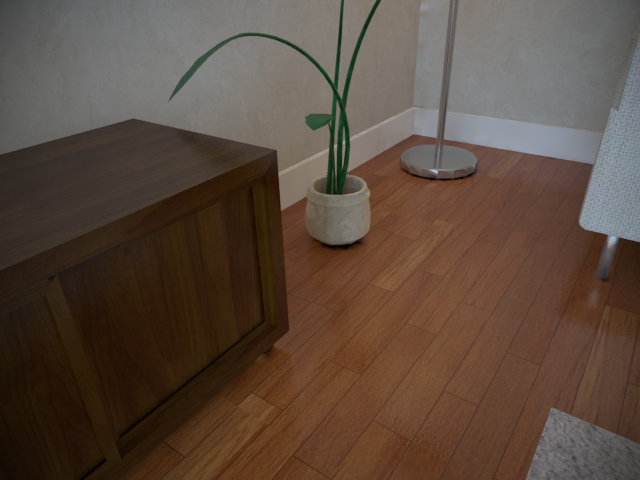}&
		\includegraphics[width=\qualitativeElemWidth{}]{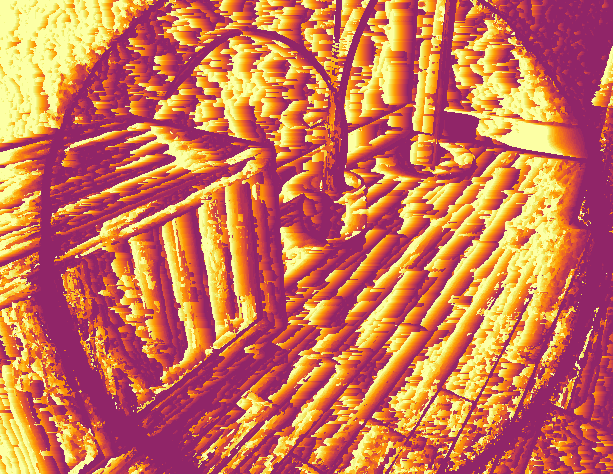}&
		\includegraphics[width=\qualitativeElemWidth{}]{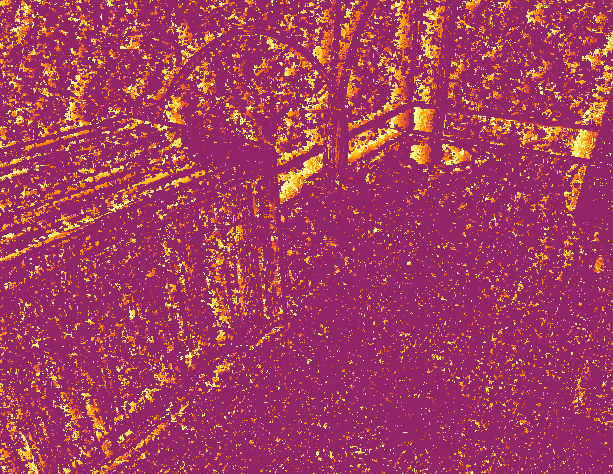}&
		\includegraphics[width=\qualitativeElemWidth{}]{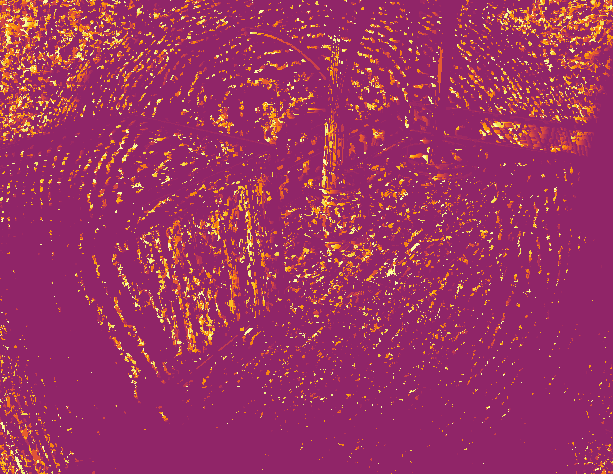}&
		\includegraphics[width=\qualitativeElemWidth{}]{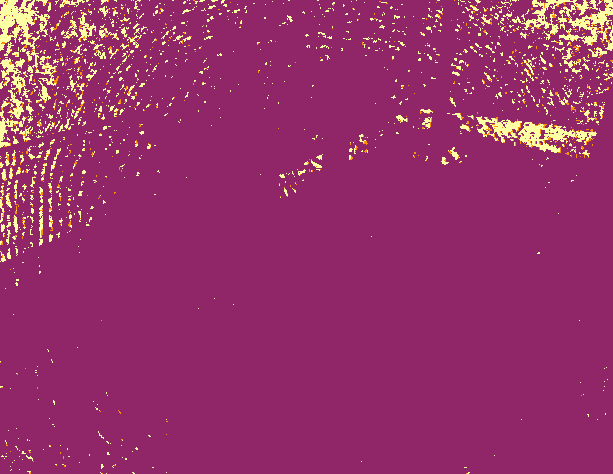}&
		\includegraphics[width=\qualitativeElemWidth{}]{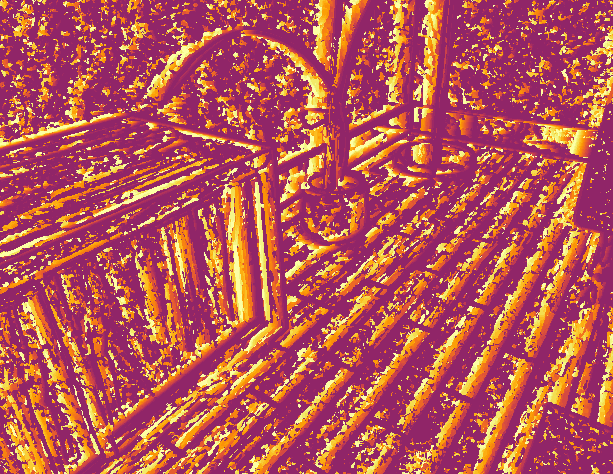}&
		\includegraphics[width=\qualitativeElemWidth{}]{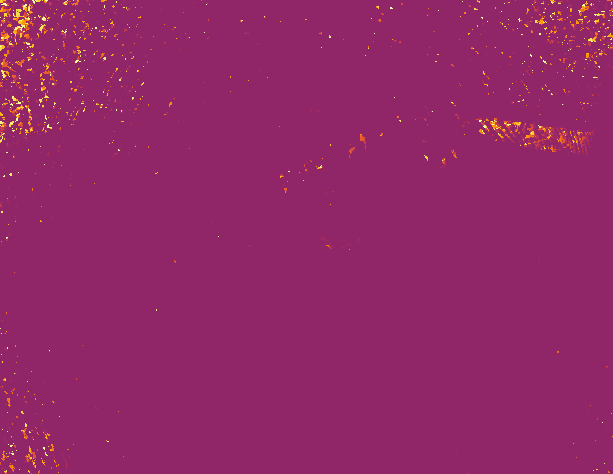} 
	\end{tabular}
	\caption{Matching cost comparison on \cite{ICLNUIM}:
		Disparity estimation against the same image with slight vignetting and different exposure time results in large disparity errors. The circle in $e_\mathit{photo}$ occurs where vignetting and exposure change cancel out. }
	\label{fig:MatchingCostSliding}
\end{figure*}
\egroup

\section{Related Work}
There has been extensive work to improve the robustness of visual odometry and visual SLAM methods towards illumination changes to ensure photometric consistency.
Typically, feature-based methods are more resilience towards illumination changes since descriptors are designed to be distinguishable even under severe changes, across different seasons and invariant of camera type. SIFT is the standard choice for Structure-from-Motion~\cite{COLMAP2016SFM} but has a significant computational cost. PTAM~\cite{PTAM} using FAST~\cite{FAST} features and ORB\_SLAM~\cite{ORBSLAM2} are two prominent examples, which show that feature-based visual SLAM can work well in many scenarios while maintaining real-time performance when exploiting binary descriptor. 

Under the assumption of a good initial guess, direct methods can obtain more accurate estimates of the camera trajectory than feature-based approaches as they exploit all intensity measurements of the images. For this reason, Dai \etal\cite{BundleFusion} use features for initialization and to constrain a subsequent dense alignment.
A popular approach, e.g. used by Schneider \etal\cite{schneider12isprs}, is to extract GoodFeaturesToTrack based on the Shi-Tomasi-Score and use the KLT optical flow tracker operating directly on intensity values. 
Similar to that, the Basalt system~\cite{Basalt} uses locally scaled intensity differences between patches at FAST features within optical flow.

A further popular method for motion estimation from camera images is LSD-SLAM~\cite{LSDSLAM}. For robustness, the authors use the Huber norm during motion estimation and map creation, while minimizing a variance-weighted photometric error. LSD-SLAM creates in parallel the map for tracking by searching along the epipolar lines minimizing the sum of squared differences. For the stereo version, Engel \etal\cite{StereoLSDSLAM} alternate between estimating a global affine function to model changing brightness and optimizing the relative pose during alignment. As an alternative, Kerl \etal\cite{kerl13icra} propose to weight the photometric residuals with a t-distribution that better matches the RGB-D sensor characteristics.

Engel \etal\cite{DSO} furthermore proposed with DSO a sparse-direct approach that further incorporates photometric calibration if available or estimates affine brightness changes with a logarithmic parametrization. They maintain an information filter to jointly estimate all involved variables.

Pascoe \etal\cite{pascoe2017nid} proposed to use the Normalized Information Distance (NID) metric for direct monocular SLAM. This works well even for tracking across seasons and under diverse illumination. Yet, the authors report to prefer photometric depth estimation for a stable initialization and only use NID after revisiting. Furthermore, Park \etal\cite{BCAVariantenPaper} presented an evaluation of different direct alignment metrics for visual SLAM. They favored the gradient magnitude due to its accuracy, robustness and speed while the census transform provided more accurate results at a much larger computational cost. In Stereo matching the census transform, e.g. in MeshStereo~\cite{MeshStereo}, and the absolute gradient difference combined with the photometric error, e.g. in StereoPatchMatch~\cite{patchmatch}, are common.

In our work\footnote{An accompanying video is available at \\{\url{https://www.ais.uni-bonn.de/videos/ICRA_2020_Gradient_Dissimilarity}}.}, we improve the gradient orientation based metric of Haber and Modersitzki~\cite{NGF} by introduction of a magnitude-dependent scaling term to simultaneously matching gradient magnitude and orientation.
We apply this to solve direct image alignment for visual odometry as well as semi-dense disparity and depth estimation. We integrated our metric in two stereo matching algorithms as well as two VO systems. Hence, we evaluate and compare the metric against existing approaches on two stereo estimation and VO datasets. 

\section{Our Method}
Our approach provides a new metric for pixel-wise matching and is easy to integrate into existing visual state estimation system. The metric measures the orientation of image gradients while also taking the magnitude into consideration. 
In the following, we denote sets and matrices with capital letters and vectors with bold lower case letters. We aim to find for a pixel $\mathbf{u}_i$ in the $i^\mathit{th}$ image the corresponding pixel $\mathbf{u}_j$ in the $j^\mathit{th}$ image that minimizes a dissimilarity measurement $e\left(\mathbf{u}_i,\mathbf{u}_j\right)$. The image coordinates $\mathbf{u}=(u_x,u_y)_F^\intercal $ are defined in the image domain $\Omega \subset \mathbb{R}^2$. For stereo matching, $i$ and $j$ correspond to the left and right image, while in direct image alignment $i$ is often the current frame and $j$ a previous (key-) frame.

A basic error function $e_\mathit{photo}$ is photometric consistency
\begin{align}
 e_\mathit{photo}\left(\mathbf{u}_i,\mathbf{u}_j\right) &= I_i\left(\mathbf{u}_i\right) - I_j\left(\mathbf{u}_j\right), \label{eq:ephoto}
\end{align}
but more robust versions often rely on intensity gradients:
\begin{align}
 e_\mathit{gm}\left(\mathbf{u}_i,\mathbf{u}_j\right) &= \left(\norm{\nabla I_i\left(\mathbf{u}_i\right)} - \norm{\nabla I_j\left(\mathbf{u}_j\right)}\right),\label{eq:emag}\\
\mathbf{e}_\mathit{gn}\left(\mathbf{u}_i,\mathbf{u}_j\right) &= \nabla I_i\left(\mathbf{u}_i\right) - \nabla I_j\left(\mathbf{u}_j\right).\label{eq:egrad}
\end{align}
The  difference of the gradients $\mathbf{e}_\mathit{gn}$ incorporates both, magnitude and orientation.  PatchMatch Stereo algorithms~\cite{patchmatch} typically combine this with the photometric error:
\begin{align}
e_\mathit{pm}(\mathbf{u}_i,\mathbf{u}_j) &= (1-\alpha) \abs{e_\mathit{photo}(\mathbf{u}_i,\mathbf{u}_j)} + \alpha \norm{e_\mathit{gn}(\mathbf{u}_i,\mathbf{u}_j)}_{\ell_1}.\label{eq:epm}
\end{align}

\newcommand\comparLeftTrim{37bp}
\newcommand\comparBottomTrim{0bp}
\newcommand\comparRightTrim{50bp}
\newcommand\comparTopTrim{0bp}
\begin{figure}[H]
	\captionsetup[subfigure]{labelformat=empty}
	
	\centering
	\setlength\tabcolsep{0.0pt} 
	\begin{tabular}{ccc}
		& 
		noise-free &
		noisy\\
		\begin{minipage}[t]{0.05\columnwidth}
			\vspace{-45pt}
			\rotatebox{90}{\small{Grayscale}}
		\end{minipage} 
		&
		\begin{minipage}[t]{0.45\columnwidth}
			\includegraphics[trim=\comparLeftTrim{} \comparBottomTrim{} \comparRightTrim{} \comparTopTrim{},clip,width=0.92\columnwidth]{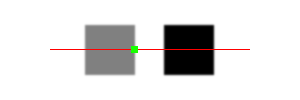}
		\end{minipage}
		&
		\begin{minipage}[t]{0.45\columnwidth}
			\includegraphics[trim=\comparLeftTrim{} \comparBottomTrim{} \comparRightTrim{} \comparTopTrim{},clip,width=0.92\columnwidth]{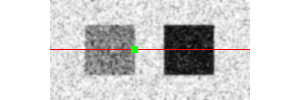}
		\end{minipage} 
		\\
		\begin{minipage}[t]{0.05\columnwidth}
			\vspace{-33pt}
			\rotatebox{90}{\small{$\nabla_\varepsilon I$}}
		\end{minipage} 
		&
		\begin{minipage}[t]{0.45\columnwidth}
			\includegraphics[trim=\comparLeftTrim{} \comparBottomTrim{} \comparRightTrim{} \comparTopTrim{},clip,width=0.92\columnwidth]{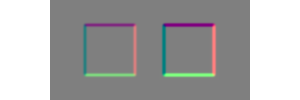}
		\end{minipage}
		&
		\begin{minipage}[t]{0.45\columnwidth}
			\includegraphics[trim=\comparLeftTrim{} \comparBottomTrim{} \comparRightTrim{} \comparTopTrim{},clip,width=0.92\columnwidth]{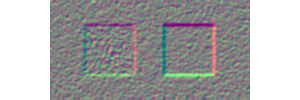}
		\end{minipage} 
		\\
		
		\rule{0pt}{0.7cm} 
		\begin{minipage}[t]{0.05\columnwidth}
			\vspace{-40pt}
			\rotatebox{90}{$e_\mathit{ugf}$}
		\end{minipage} 
		&
		\begin{minipage}[t]{0.45\columnwidth}
			\begin{flushleft}
				\begin{tikzpicture}
				\noindent
				\begin{axis}[
				enlargelimits=false,
				width=4.9cm,
				height=3cm,
				ymax=1,
				y tick label style={/pgf/number format/.cd, fixed, fixed zerofill,precision=1},
				yticklabels={,,},
				xticklabels={,,}
				]
				\addplot [color=blue]  table [x=x_val, y=matching_cost, col sep=comma, mark=none,] {./data/ugf_cost.csv};
				\end{axis}

				\end{tikzpicture}
			\end{flushleft}
		\end{minipage} 
		&
		\begin{minipage}[t]{0.45\columnwidth}
			\begin{flushleft}
				\begin{tikzpicture}
				\noindent
				\begin{axis}[
				enlargelimits=false,
				width=4.9cm,
				height=3cm,
				ymax=1,
				y tick label style={/pgf/number format/.cd, fixed, fixed zerofill,precision=1},
				yticklabels={,,},
				xticklabels={,,}
				]
				\addplot [color=blue]  table [x=x_val, y=matching_cost, col sep=comma, mark=none,
				] {./data/ugf_cost_noisy.csv};
				\end{axis}
				\end{tikzpicture}
			\end{flushleft}
		\end{minipage} 
		\\
		
		\begin{minipage}[t]{0.05\columnwidth}
			\vspace{-5pt}
			\rotatebox{90}{$e_\mathit{mag}$}
		\end{minipage} 
		&
		\begin{minipage}[t]{0.45\columnwidth}
			\vspace{-14pt}
			\begin{flushleft}
				\begin{tikzpicture}
				\noindent
				\begin{axis}[
				enlargelimits=false,
				width=4.9cm,
				height=3cm,
				ymax=0.217,
				xmin=0,
				xmax=200,
				y tick label style={/pgf/number format/.cd, fixed, fixed zerofill,precision=1},
				yticklabels={,,},
				xticklabels={,,}
				]				
				\addplot [color=blue]  table [x=x_val, y=matching_cost, col sep=comma, mark=none,
				] {./data/grad_mag.csv};
				\end{axis}
				\end{tikzpicture}
			\end{flushleft}
		\end{minipage} 
		&
		\begin{minipage}[t]{0.45\columnwidth}
			\vspace{-14pt}
			\begin{flushleft}
				\begin{tikzpicture}
				\noindent
				\begin{axis}[
				enlargelimits=false,
				width=4.9cm,
				height=3cm,
				ymax=0.17,
				xmin=0,
				xmax=200,
				y tick label style={/pgf/number format/.cd, fixed, fixed zerofill,precision=1},
				yticklabels={,,},
				xticklabels={,,}
				]
				\addplot [color=blue, mark minGreen]  table [x=x_val, y=matching_cost, col sep=comma, mark=none,
				] {./data/grad_mag_noisy.csv};
				\end{axis}
				\end{tikzpicture}
			\end{flushleft}
		\end{minipage} 
		\\
		
		\begin{minipage}[t]{0.05\columnwidth}
			\vspace{-0pt}
			\rotatebox{90}{$e_\mathit{sgf}$}
		\end{minipage} 
		&
		\begin{minipage}[t]{0.45\columnwidth}
			\vspace{-14pt}
			\begin{flushleft}
				\begin{tikzpicture}
				\noindent
				\begin{axis}[
				enlargelimits=false,
				width=4.9cm,
				height=3cm,
				ymax=1,
				y tick label style={/pgf/number format/.cd, fixed, fixed zerofill,precision=1},
				yticklabels={,,},
				xticklabels={,,}
				]				
				\addplot [color=blue, mark minGreen]  table [x=x_val, y=matching_cost, col sep=comma, mark=none,
				] {./data/sgf_cost.csv};				
				\end{axis}
				\end{tikzpicture}
			\end{flushleft}
		\end{minipage} 
		&
		\begin{minipage}[t]{0.45\columnwidth}
			\vspace{-14pt}
			\begin{flushleft}
				\begin{tikzpicture}
				\noindent
				\begin{axis}[
				enlargelimits=false,
				width=4.9cm,
				height=3cm,
				ymax=1,
				y tick label style={/pgf/number format/.cd, fixed, fixed zerofill,precision=1},
				yticklabels={,,},
				xticklabels={,,}
				]
				\addplot [color=blue, mark minGreen]  table [x=x_val, y=matching_cost, col sep=comma, mark=none,
				] {./data/sgf_cost_noisy.csv};
				\end{axis}
				\end{tikzpicture}
			\end{flushleft}
		\end{minipage} 
		\\
		
		\begin{minipage}[t]{0.05\columnwidth}
			\vspace{-5pt}
			\rotatebox{90}{$e_\mathit{photo}$}
		\end{minipage} 
		&
		\begin{minipage}[t]{0.45\columnwidth}
			\vspace{-11pt}
			\begin{flushleft}
				\begin{tikzpicture}
				\noindent
				\begin{axis}[
				enlargelimits=false,
				width=4.9cm,
				height=3cm,
				ymax=0.6503,
				xmin=0,
				xmax=200,
				y tick label style={/pgf/number format/.cd, fixed, fixed zerofill,precision=1},
				yticklabels={,,},
				xticklabels={,,}
				]
				\addplot [color=blue]  table [x=x_val, y=matching_cost, col sep=comma, mark=none,
				] {./data/bca.csv};				
				\end{axis}
				\end{tikzpicture}
			\end{flushleft}
		\end{minipage} 
		&
		\begin{minipage}[t]{0.45\columnwidth}
			\vspace{-11pt}
			\begin{flushleft}
				\begin{tikzpicture}
				\noindent
				\begin{axis}[
				enlargelimits=false,
				width=4.9cm,
				height=3cm,
				ymax=0.48,
				xmin=0,
				xmax=200,
				y tick label style={/pgf/number format/.cd, fixed, fixed zerofill,precision=1},
				yticklabels={,,},
				xticklabels={,,}
				]
				\addplot [color=blue]  table [x=x_val, y=matching_cost, col sep=comma, mark=none,
				] {./data/bca_noisy.csv};
				\end{axis}
				\end{tikzpicture}
			\end{flushleft}
		\end{minipage} 	
		\vspace{-8pt}		
	\end{tabular}
	
	\caption{Error comparison for gradient based metrics on a toy example. The lower boxes show the error between the green reference box and a shifted box along the red horizontal line. $e_\mathit{ugf}$ prefers strong edges with same orientation, while $e_\mathit{mag}$ does not take the orientation into account and thus generates further local minima. Our $e_\mathit{sgf}$ provides the correct minima which are marked with a green circle.}\label{fig:Metric}
\end{figure}
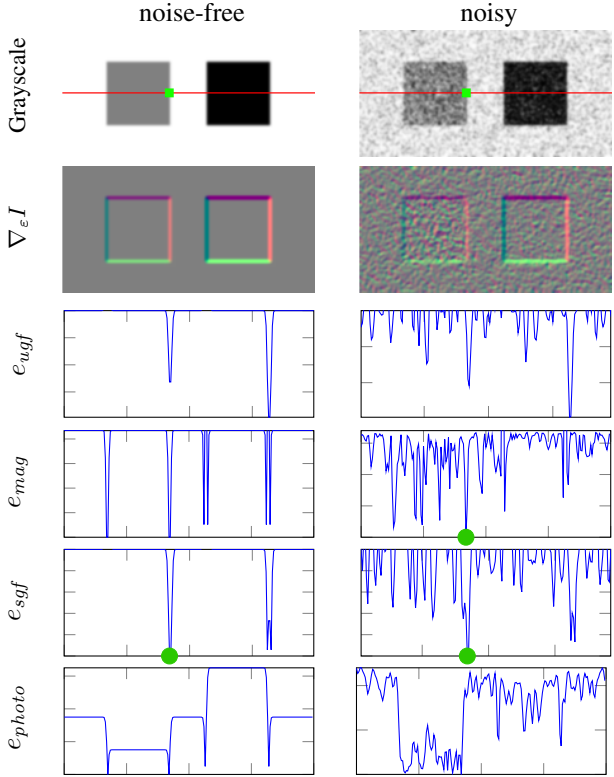

\subsection{Normalized Gradient-based Direct Image Alignment}
A complementary approach is to align the gradients orientation. The na\"ive approach may use the costly atan-operation to obtain the orientation angle $\theta$ and simply calculate differences. Instead, we follow the approach of \cite{NGF,taylor2015gom} to use the dot product and its relation to the cosine as a measure of orientation. If the two vectors $\mathbf{a},\mathbf{b}$ have unit length, the dot product is equal to the cosine of the angle between the vectors, which is zero for perpendicular vectors, one for same  and minus one for opposite orientation.
Simply normalizing the gradient by its magnitude is undesirable as noise in low gradient regions will predominate the orientation. Hence, Taylor \etal\cite{taylor2015gom} normalizes the dot product by its magnitude over a window:
\begin{align}
e_\mathit{gom}\left(\mathbf{u}_i,\mathbf{u}_j\right) &= 1-\frac{\sum_{u\in W} \lvert\nabla I_i\left(\mathbf{u}_i \right)\cdot\nabla I_j\left(\mathbf{u}_j\right)\rvert}{\sum_{u\in W} \norm{\nabla I_i\left(\mathbf{u}_i\right)}\norm{\nabla I_j\left(\mathbf{u}_j\right)}}.
\end{align}
Instead we follow \cite{NGF} and regularize the magnitude by a parameter $\varepsilon$:
\begin{align}
\varepsilon &= \frac{1}{\lvert\Omega\lvert} \sum_{\mathbf{u}\in\Omega} \norm{\nabla I(\mathbf{u})}^2,\\
\nabla_\varepsilon I &= \frac{\nabla I}{\sqrt{\norm{\nabla I}^2+\varepsilon}}.
\end{align}
This effectively downweighs the gradients magnitude in low gradient regions such that $\norm{\nabla_\varepsilon I}$ will be close to zero. We estimate the parameter $\varepsilon$ on a per image basis and will use $\varepsilon$ and $\vartheta$ to make the distinction between different images more visible.

\begin{figure}[]
	\captionsetup[subfloat]{labelformat=empty}
	\centering
	
	\subfloat[a) $e_\mathit{ugf}$]{
		\includegraphics[ width=0.45\columnwidth] {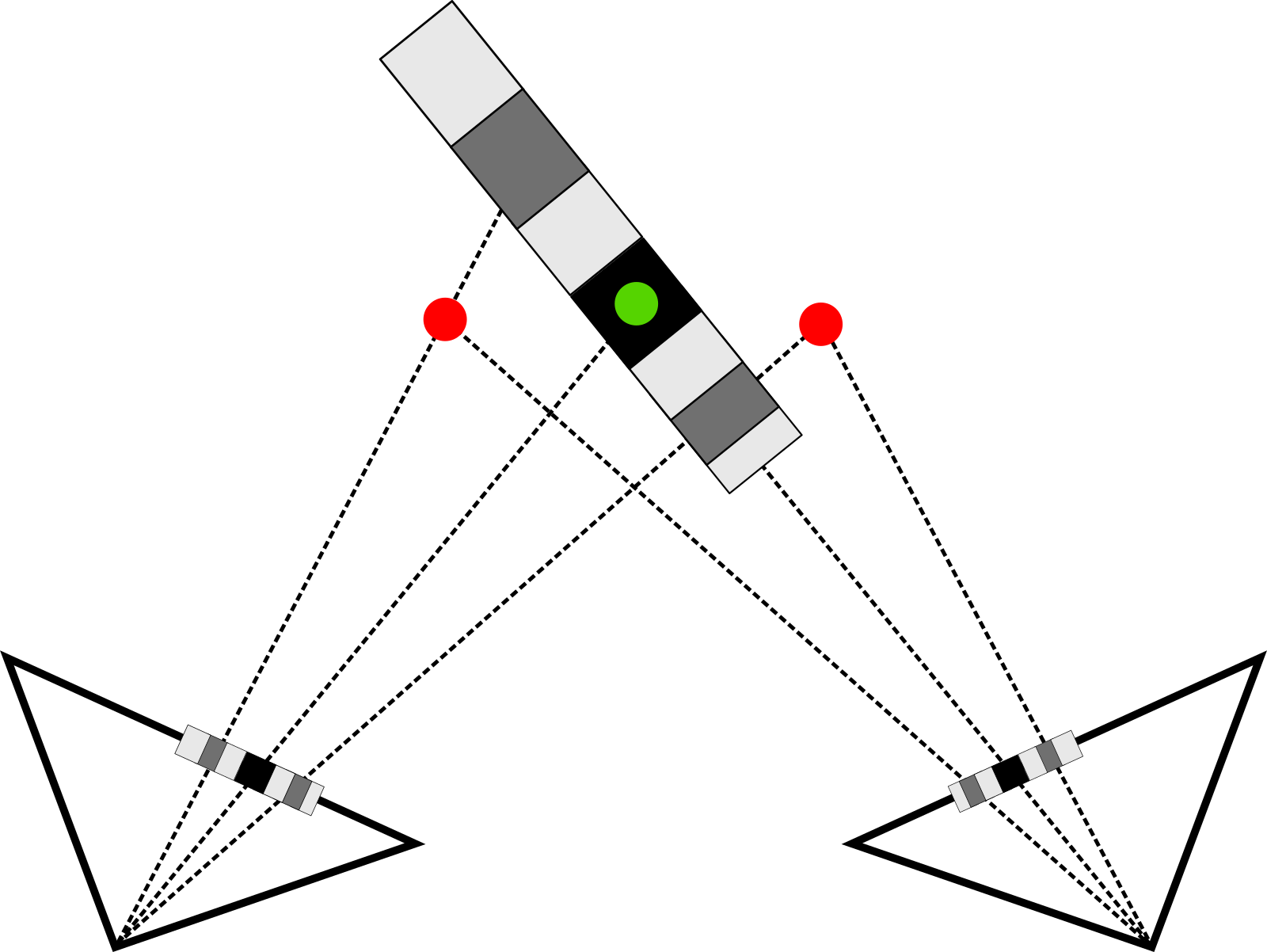}
	}
	\subfloat[b) $e_\mathit{sgf}$]{
		\includegraphics[ width=0.45\columnwidth] {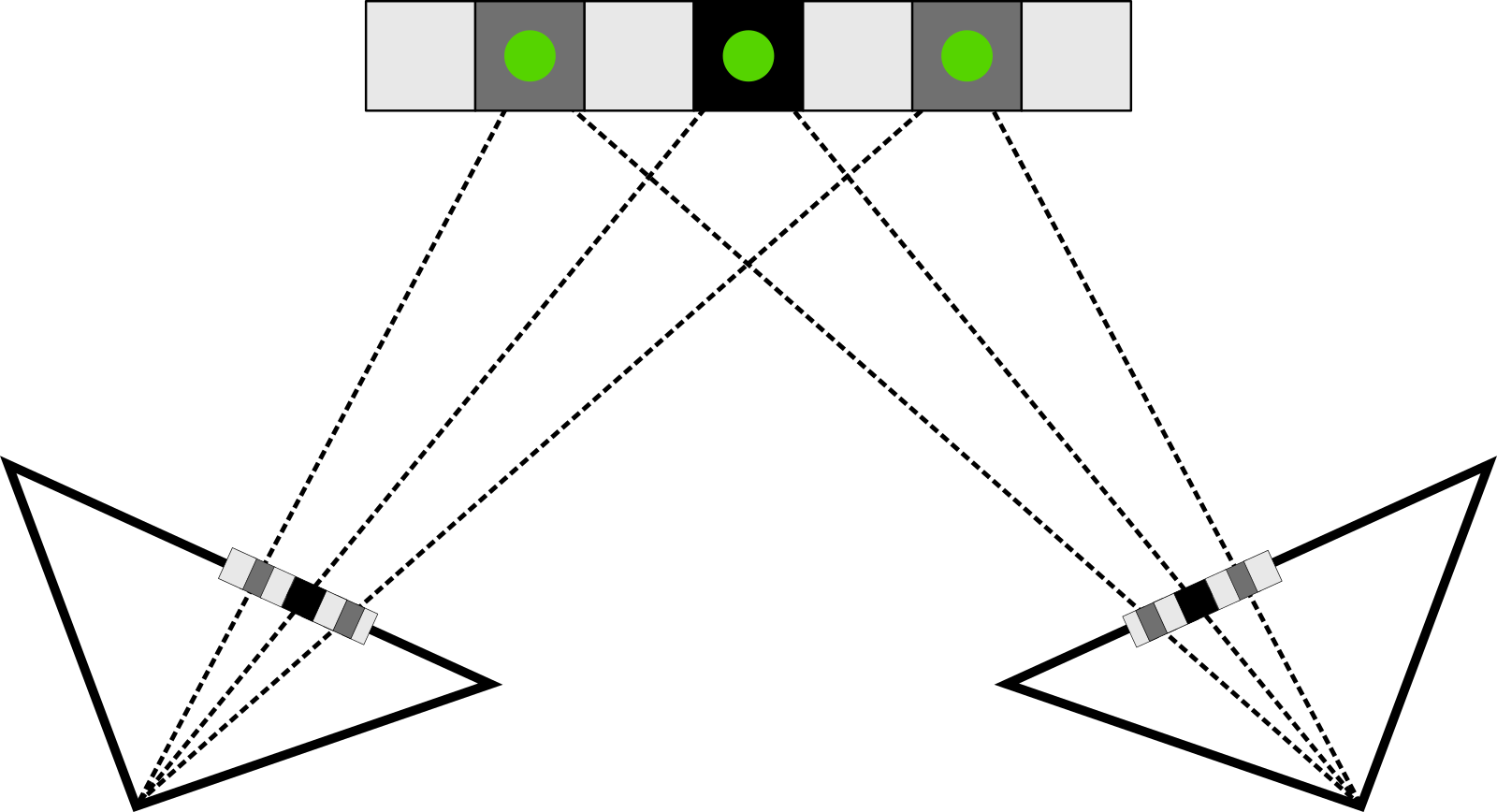}
	}
	\caption{Association impact: $e_\mathit{ngf}$ and $e_\mathit{ugf}$ tend to match patches with similar gradient orientation but stronger magnitude. This can cause severe distortions in the 3D reconstruction (left). Associating patches with similar gradient orientation and magnitude using $e_\mathit{sgf}$ allows for correct triangulation (right).}
	\label{fig:Assoc}
\end{figure}

In the context of multi-modal image registration the authors of \cite{NGF} minimize the per pixel error $e_\mathit{ngf}$:
\begin{align}
 e_\mathit{ngf}\left(\mathbf{u}_i,\mathbf{u}_j\right) &= 1 - \left[\nabla_\varepsilon I_i(\mathbf{u}_i) \cdot \nabla_\vartheta I_j(\mathbf{u}_j)\right]^2.\label{eq:engf}
\end{align}
Squaring the dot product, or taking the absolute value, ensures, that not only gradients with same orientation but also with opposite orientation coincide.
This is important for registering CT to MRT data and vice versa where the image gradients may have opposite direction. This error has an important flaw as low gradient pixels prefer to match with higher magnitude ones rather than similar gradients. If the largest magnitude edge is always matched, we would obtain inconsistent depth estimates with high reprojection errors or when successively reducing the search region skew the region and obtain wrong estimates as visualized in \reffig{fig:Assoc}.

Since we want to use images from the same sensor type, we can omit the square and only use the following residual:
\begin{align}
e_\mathit{ugf}(\mathbf{u}_i,\mathbf{u}_j) &= 1 - \nabla_\vartheta I_j\left(\mathbf{u}_j\right) \cdot \nabla_\varepsilon I_i\left(\mathbf{u}_i\right).
 \end{align}
The errors $e_\mathit{ngf}$ and $e_\mathit{ugf}$ are bounded in the interval $[0,2]$. To ensure the correct behavior for smaller gradients as visualized in \reffig{fig:Metric}, we scale the dot product by the maximum value: 
\begin{align}
e_\mathit{sgf}(\mathbf{u}_i,\mathbf{u}_j) &= 1 - \frac{\nabla_\vartheta I_j\left(\mathbf{u}_j\right) \cdot \nabla_\varepsilon I_i\left(\mathbf{u}_i\right)}{\max\left(\norm{\nabla_\varepsilon I_i\left(\mathbf{u}_i\right)}^2, \norm{\nabla_\vartheta I_j\left(\mathbf{u}_j\right)}^2,\tau\right)}.
\end{align}
The scaling term of SGF thereby increases the number of successfully estimated points in semi-dense depth estimation. Here, $\tau$ is a small constant to prevent division by zero.

To further reduce the number of mathematical operations in above equation, especially the division by the regularized norm, we derived two further combinations of orientation and magnitude:
\begin{align}
n\left(\mathbf{u}_i,\mathbf{u}_j\right) &= \nabla I_j\left(\mathbf{u}_j\right) \cdot \nabla I_i\left(\mathbf{u}_i\right),\\
 nij(\mathbf{u}_i,\mathbf{u}_j) &= \frac{\norm{\nabla_\vartheta I_j\left(\mathbf{u}_j\right)}}{\norm{\nabla_\varepsilon I_i\left(\mathbf{u}_i\right)}}\norm{\nabla I_i\left(\mathbf{u}_i\right)}^2,\\
 nji(\mathbf{u}_i,\mathbf{u}_j) &= \frac{\norm{\nabla_\varepsilon I_i\left(\mathbf{u}_i\right)}}{\norm{\nabla_\vartheta I_j\left(\mathbf{u}_j\right)}}\norm{\nabla I_j\left(\mathbf{u}_j\right)}^2,\\
 e_\mathit{sgf2}(\mathbf{u}_i,\mathbf{u}_j) &= \max\left(nij,nji\right) - n\left(\mathbf{u}_i,\mathbf{u}_j\right)\\
 e_\mathit{sgf3}(\mathbf{u}_i,\mathbf{u}_j) &= \norm{\nabla I_i\left(\mathbf{u}_i\right)}\norm{\nabla I_j\left(\mathbf{u}_j\right)} - n\left(\mathbf{u}_i,\mathbf{u}_j\right).
\end{align}

Given a formulation for the error, we can now formulate stereo matching and direct image alignment. The former aims to find for each pixel $\mathbf{u}_l$ in the left image the corresponding pixel $\mathbf{u}_r$ in the right image that minimizes a dissimilarity measurement $e\left(\mathbf{u}_l,\mathbf{u}_r\right)$:
\begin{align}
d^\ast_\mathbf{u} & = \argmin_{d\in\mathcal{R}} \sum_{\mathbf{u}_l\in W} e\left(\mathbf{u}_l,\mathbf{u}_r\left(d\right)\right),\\
\mathbf{u}_r\left(d\right) &= \mathbf{u}_l - \left(d,0\right)^\intercal.
\end{align}
Here, the disparity $d$ is defined as the distance along the x-axis of the stereo rectified left and right image pair.
For robustness, the error function $e$ is calculated over a patch $W_\mathbf{u}$ with window size $w$ centered around the pixel $\mathbf{u}$ rather than a single pixel. In the latter, we seek the transformation $T_{cr}$ that aligns the reference with the current image optimally w.r.t. an error metric $e$ between a reference pixel-patch $\mathcal{N}_{\mathbf{p}_r}$ around $\mathbf{p}_r$ and its projection onto $I_c$:
\begin{align}
  T_{cr} = \argmin \sum_{\mathbf{p}_r \in \mathcal{M}} \sum_{\mathbf{p}_k \in \mathcal{N}_{\mathbf{p}_r}} \rho\left(\norm{e\left(\mathbf{p}_i\right)}^2\right)\label{eq:opt}.
\end{align}
A robust cost function $\rho$ like the Huber norm reduces the effect of outliers. This minimization is typically solved iteratively with the standard Gauss-Newton algorithm.

Hence, the Jacobian for $e_\mathit{sgf}$ w.r.t. the pixel $\mathbf{u}_i$ is needed:
\begin{align}
nn &= \nabla_\vartheta I_j\left(\mathbf{u}_j\right) \cdot \nabla_\varepsilon I_i\left(\mathbf{u}_i\right),\\
s_1 &= nn \begin{cases}
-1, & \text{if } \scriptscriptstyle{\norm{\nabla_\vartheta I_j}^2 > \norm{\nabla_\varepsilon I_i}^2}\\
1 - \frac{2}{\norm{\nabla_\varepsilon I_i}}, &\text{otherwise}
\end{cases}\\
\frac{\partial e_\mathit{sgf}}{\partial \mathbf{u}_i} &= -\frac{\left(\nabla_\vartheta I_j+ s_1 \nabla_\varepsilon I_i\right)^\intercal}{\max\left(\norm{\nabla_\varepsilon I_i}, \norm{\nabla_\vartheta I_j}\right)} \frac{\left(\nabla_2\right)I_i}{\norm{\nabla I_i}_\varepsilon},\\[10pt]
s_2 &= \begin{cases}
\frac{\norm{\nabla_\vartheta I_j}}{\norm{\nabla_\varepsilon I_i}}\left(2-\frac{\norm{\nabla I_i}^2}{\norm{\nabla I_i}^2+\varepsilon}\right), & \text{if } \scriptstyle{nij > nji}\\
\frac{\norm{\nabla_\varepsilon I_i}}{\norm{\nabla_\vartheta I_j}} \frac{\norm{\nabla I_j}^2}{\left(\norm{\nabla I_i}^2+\varepsilon\right)}, &\text{otherwise}
\end{cases}\\
\frac{\partial e_\mathit{sgf2}}{\partial \mathbf{u}_i} &= \left(s_2\nabla I_i - \nabla I_j\right)\left(\nabla_2\right)I_i,\\[10pt]
\frac{\partial e_\mathit{sgf3}}{\partial \mathbf{u}_i} &= \left(\frac{1}{2}\frac{\norm{\nabla I_j}}{\norm{\nabla I_i}}\nabla I_i - \nabla I_j\right)\left(\nabla_2\right)I_i.
\end{align}
Here, $\left(\nabla_2\right) I_i$ denotes the hessian of the intensity at pixel $\mathbf{u}_i$. 

\bgroup
\newcolumntype{Y}{>{\centering\arraybackslash}X}
\renewcommand{\arraystretch}{1.3}
\begin{table}[]
\centering
\caption{Evaluation on Middlebury Stereo 2014 training set~\cite{msb2014}}
\begin{tabularx}{\columnwidth}{@{\extracolsep{\fill}}|l|l|Y|Y|Y|Y|Y|Y}
\cline{3-7}
\multicolumn{1}{c}{} & & Orig. & $e_\mathit{sad}$ & $e_\mathit{agm}$ & $e_\mathit{pm}$ & $e_\mathit{sgf}$ \\\hline
\multirow{5}{*}{\rotatebox{90}{StereoBM}}
 & mean & 7.20 & 5.80 & 6.31 & 4.56 & \textbf{3.29} \\
 & bad 1 & 18.36 & 20.51 & 21.33 & 17.19 & \textbf{12.60} \\
 & bad 2 & 16.41 & 17.01 & 17.79 & 14.25 & \textbf{10.36} \\
 & bad 4 & 14.88 & 14.19 & 14.69 & 11.94 & \textbf{8.61} \\
 & invalid & 40.44 & \textbf{34.51} & 52.69 & 44.74 & 45.49 \\\hline
 
\multirow{5}{*}{\rotatebox{90}{MeshStereo}}
 & mean & 5.68 & 11.22 & 7.85 & 6.70 & \textbf{4.17} \\
 & bad 1 & \textbf{16.87} & 46.55 & 33.45 & 28.51 & 20.61 \\
 & bad 2 & \textbf{13.02} & 40.25 & 27.38 & 23.32 & 15.94 \\
 & bad 4 & \textbf{10.71} & 33.18 & 22.02 & 18.78 & 12.53 \\
 & invalid & \textbf{0.01} & 1.01 & 0.09 & 0.08 & 0.04 \\\hline
\end{tabularx}
\label{tab:MSB}
\end{table}
\egroup

\bgroup
\newcolumntype{Y}{>{\centering\arraybackslash}X}
\renewcommand{\arraystretch}{1.3} 
\begin{table}[]
\centering
\caption{Evaluation on KITTI Stereo 2015 training set~\cite{kitti2015}}
\begin{tabularx}{\columnwidth}{@{\extracolsep{\fill}}|l|l|Y|Y|Y|Y|Y|Y}
\cline{3-7}
\multicolumn{1}{c}{}&  & Orig. & $e_\mathit{sad}$ & $e_\mathit{agm}$ & $e_\mathit{pm}$ & $e_\mathit{sgf}$ \\\hline
\multirow{5}{*}{\rotatebox{90}{StereoBM}}
 & mean & 6.11 & 3.21 & 3.17 & 1.74 & \textbf{1.61} \\
 & bad 1 & 19.80 & 19.79 & 22.13 & 15.93 & \textbf{13.99} \\
 & bad 2 & 11.60 & 10.07 & 11.04 & 6.87 & \textbf{5.91} \\
 & bad 4 & 9.03 & 6.34 & 6.73 & 3.94 & \textbf{3.41} \\
 & invalid & 46.74 & \textbf{29.57} & 53.02 & 39.33 & 45.17 \\\hline
 
\multirow{5}{*}{\rotatebox{90}{MeshStereo}}
 & mean & 2.03 & 2.94 & 2.92 & 2.07 & \textbf{2.02} \\
 & bad 1 & \textbf{27.95} & 42.34 & 33.84 & 29.60 & 29.35 \\
 & bad 2 & \textbf{12.00} & 25.45 & 17.32 & 13.67 & 13.48 \\
 & bad 4 & \textbf{5.57} & 14.01 & 8.85 & 6.77 & 6.67 \\
 & invalid & 0.07 & 0.15 & 0.10 & 0.08 & \textbf{0.06}\\\hline
 \end{tabularx}
\label{tab:kitti}
\end{table}
\egroup

\bgroup
\newcolumntype{Y}{>{\centering\arraybackslash}X}
\renewcommand{\arraystretch}{1.3}
\newcolumntype{?}{!{\vrule width 1pt}}
\newcolumntype{C}[1]{>{\centering\arraybackslash\hspace{0pt}}m{#1}}
\begin{table*}[]
	\centering
	\caption{ATE results in meters on EuRoC dataset~\cite{eurocMAV}.
	}
	\begin{tabularx}{\textwidth}{ | C{0.3cm} | C{2cm} | @{\extracolsep{\fill}} |Y|Y|Y|Y|Y|Y|Y|Y|Y|Y?Y|}
		\cline{3-13}
\multicolumn{2}{c|}{} & MH1 & MH2 & MH3 & MH4 & MH5 & V11 & V12 & V13 & V21 & V22 & Avg \\
		\hline
\multirow{5}{*}{\rotatebox{90}{Original}} & OKVIS & 0.085 & 0.083 & 0.135 & 0.143 & 0.278 & 0.041 & 0.956 & 0.102 & 0.054 & 0.063 & 0.194\\
		\cline{2-13}
& ORB-SLAM2 & 0.124 & 0.094 & 0.253 & 0.151 & 0.132 & 0.090 & 0.219 & 0.270 & 0.149 & 0.203 & 0.168\\
		\cline{2-13}
& SVO2 & 0.093 & 0.111 & 0.355 & 2.444 & 0.456 & 0.074 & 0.174 & 0.270 & 0.109 & 0.158 & 0.424\\	
		\cline{2-13}
& DSO & \textbf{0.051} & 0.045 & 0.165 & 0.164 & 0.460 & 0.194 & 0.151 & 1.075 & 0.080 & 0.098 & 0.227\\
		\cline{2-13}
& Basalt & 0.076 & 0.045 & 0.058 & 0.096 & 0.141 & 0.041 & 0.052 & 0.073 & 0.032 & \textbf{0.046} & 0.066\\
		\hline
		\hline
\multirow{7}{*}{\rotatebox{90}{Ours}} & DSO w/ $e_\mathit{sgf}$ & 0.071 & 0.050 &    0.264 & 0.235 & 0.237 & 0.142 & 0.178 & 0.933 & 0.072 & 0.086 & 0.206 \\
		\cline{2-13}
& Basalt w/ $e_\mathit{gm}$ & 0.090 & 0.044 & 0.084 & \textbf{0.091} & 0.135 & 0.049 & 0.099 &    0.161 & 0.030 & 0.079 & 0.086 \\
		\cline{2-13}
& Basalt w/ $\mathbf{e}_\mathit{gn}$ & 0.076 & 0.055 & 0.057 & 0.112 & 0.115 & \textbf{0.039} & \textbf{0.042} &   0.093 & 0.037 & 0.048 & 0.067\\
		\cline{2-13}
& Basalt w/ $e_\mathit{sgf}$ & 0.078 & 0.062 & 0.080 & 0.215 & 0.111 & 0.043 & 0.107 & 0.156 & 0.037 & 0.108 & 0.100 \\
		\cline{2-13}
& Basalt w/ $e_\mathit{sgf2}$ & 0.086 & 0.065 & 0.081 & 0.109 & 0.148 & 0.040 & 0.069 & \textbf{0.061} & \textbf{0.029} & 0.058 & 0.075 \\
		\cline{2-13}
& Basalt w/ $e_\mathit{sgf3}$ & 0.061 & \textbf{0.042} & \textbf{0.065} & 0.094 & \textbf{0.106} & 0.041 & 0.056  & 0.082 & 0.034 & 0.054 & \textbf{0.063} \\
		\hline
	\end{tabularx}
	\label{tab:EurocATE}
\end{table*}
\egroup

\section{Evaluation}

The first experiment is designed to illustrate the  robustness of our metric under small image variations.
To underline how even minimal image variations impact the dissimilarity metrics, we used images 
from the ICL-NUIM "lr kt2" sequence~\cite{ICLNUIM} and changed the exposure time and added a vignetting to frames \num{120} and \num{808}, see \reffig{fig:MatchingCostSliding} for a visualization. The disparity error is minimal in green regions with ideal disparity being \num{0} and window size \num{3}. We evaluated $d\in[0,20)$ for the different metrics. 
As expected, $e_\mathit{photo}$ is large (avg. \num{8.13}~px / \num{7.76}~px ), while gradient orientation alone ($e_\mathit{ugf}$) achieves on avg. \num{4.49}~px / \num{4.78}~px. Normalized cross-correlation ($e_\mathit{ncc}$) results in a disparity error of \num{3.04}~px / \num{2.38}~px. The magnitude ($e_\mathit{mag}$) is better suited (\num{2.11}~px / \num{1.40}~px) while $e_\mathit{gom}$ (\num{2.02}~px / \num{0.49}~px) and $e_\mathit{pm}$(\num{1.24}~px / \num{0.18}~px) perform best after our metric (\num{1.21}~px / \num{0.18}~px) showing the smallest dissimilarity values.

\bgroup
\newcommand{\qualitativeElemWidth}{3.9cm}
\renewcommand{\tabcolsep}{1pt}
\begin{figure}[h!]
	\centering
	\begin{tabular}{lccl}
		\parbox[t]{3mm}{\rotatebox[origin=c]{90}{\small RGB}} & \includegraphics[width=\qualitativeElemWidth{},align=c]{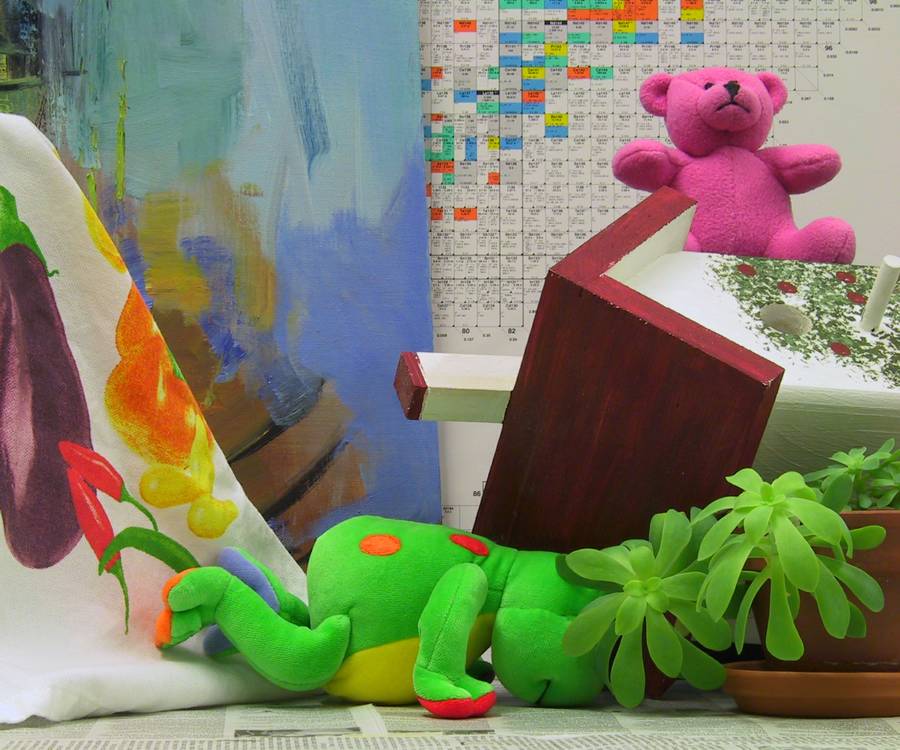} & \includegraphics[width=\qualitativeElemWidth{},align=c]{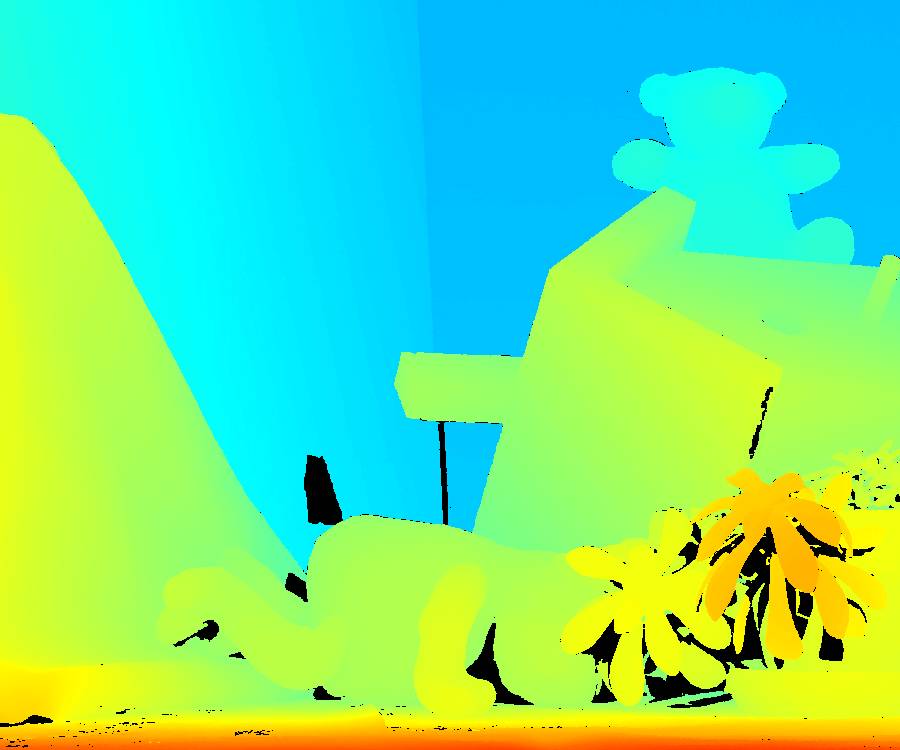} & \parbox[t]{3mm}{\rotatebox[origin=c]{90}{\small GT}} \vspace{0.5mm} \\
		\parbox[t]{3mm}{\rotatebox[origin=c]{90}{\small Orig.}} & \includegraphics[width=\qualitativeElemWidth{},align=c]{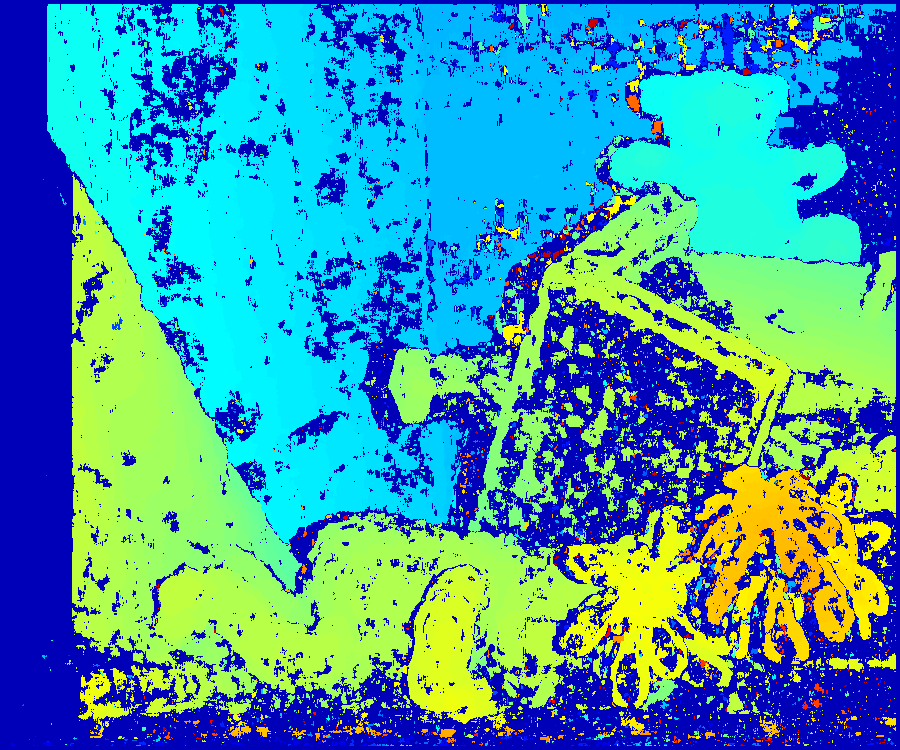} & \includegraphics[width=\qualitativeElemWidth{},align=c]{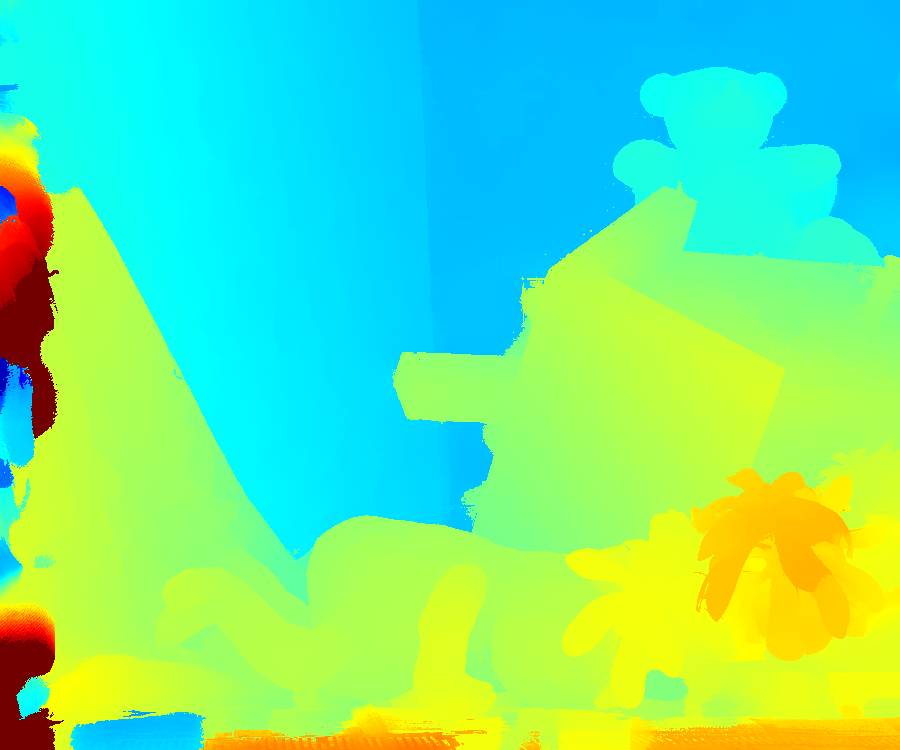} & \parbox[t]{3mm}{\rotatebox[origin=c]{90}{\small Orig.}} \vspace{0.5mm} \\
		\parbox[t]{3mm}{\rotatebox[origin=c]{90}{\small $e_\mathit{pm}$}} &  \includegraphics[width=\qualitativeElemWidth{},align=c]{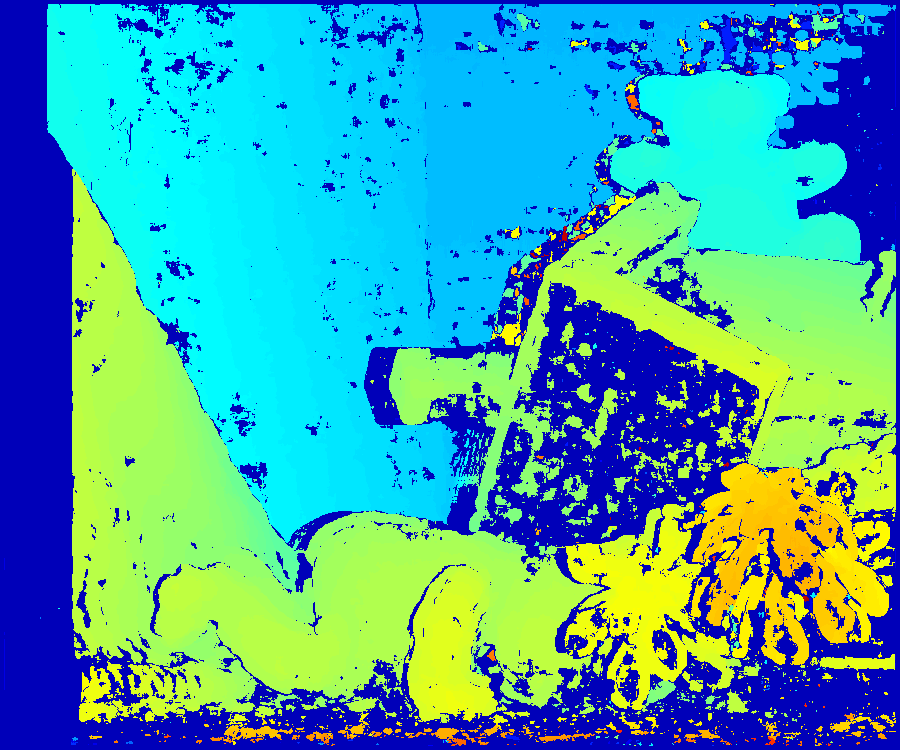} & \includegraphics[width=\qualitativeElemWidth{},align=c]{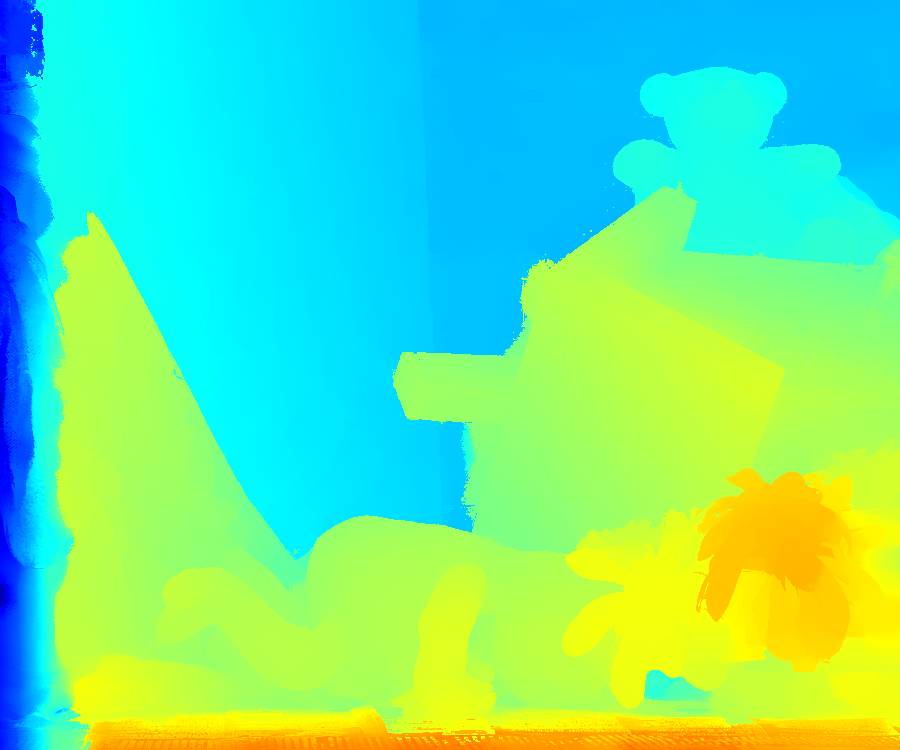} & \parbox[t]{3mm}{\rotatebox[origin=c]{90}{\small $e_\mathit{pm}$}} \vspace{0.5mm} \\
		\parbox[t]{3mm}{\rotatebox[origin=c]{90}{\small $e_\mathit{sgf}$}} & \includegraphics[width=\qualitativeElemWidth{},align=c]{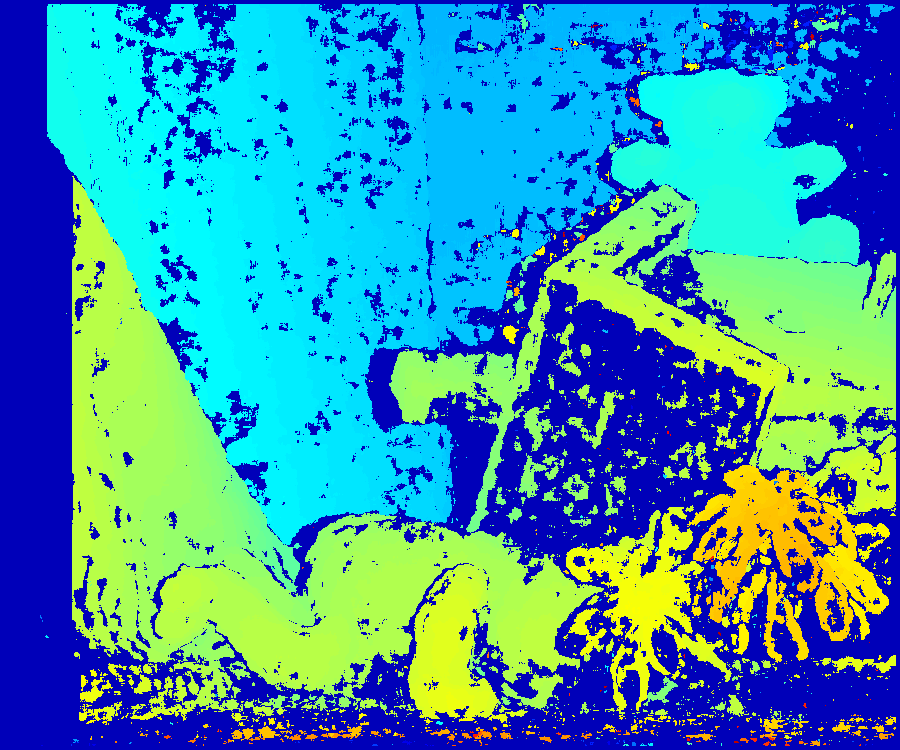} & \includegraphics[width=\qualitativeElemWidth{},align=c]{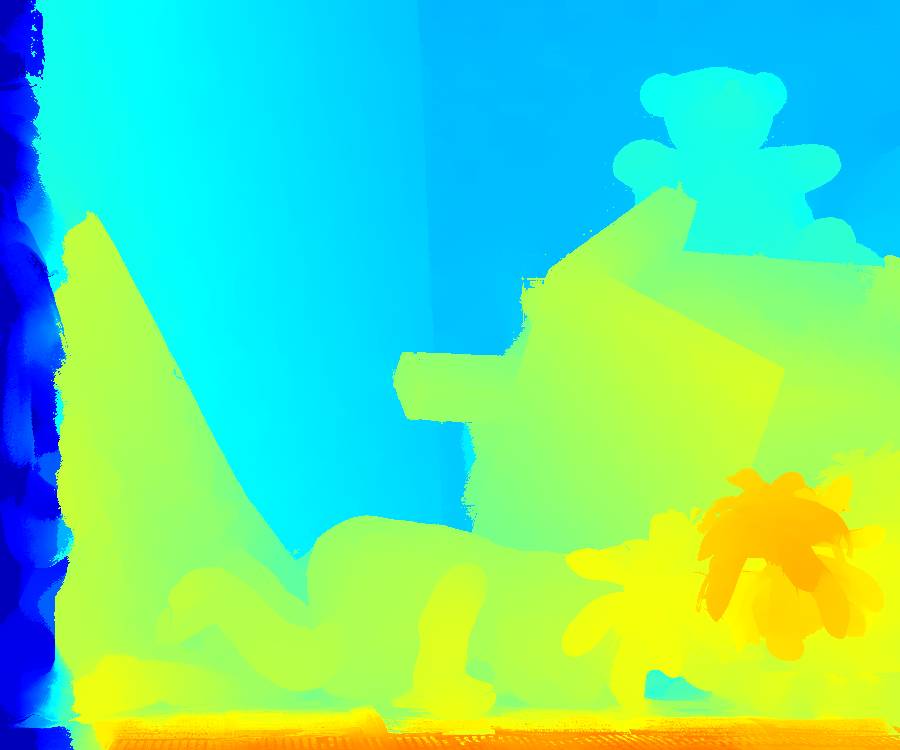} & \parbox[t]{3mm}{\rotatebox[origin=c]{90}{\small $e_\mathit{sgf}$}}\\
		& {\small StereoBM} & {\small MeshStereo} & 
	\end{tabular}
	\caption{Disparity comparison on Teddy of the Middlebury Stereo 2014 Benchmark~\cite{msb2014} for the original algorithms and the two best metrics.} 
	\label{fig:Msb}
\end{figure}
\egroup

The second experiment is designed to show our metrics suitability for (semi-) dense depth estimation supporting the first claim. For this, we integrated a variety of metrics for cost volume calculation into OpenCVs stereo block matching as well as the more sophisticated MeshStereo algorithm~\cite{MeshStereo}. We evaluate the mean disparity error and report the percentage of bad pixels with \num{1}, \num{2}, and \num{4}~px disparity error. Both algorithms are tested on the training sets of the Middlebury Stereo Benchmark~\cite{msb2014} (half size) and the KITTI Stereo Benchmark~\cite{kitti2015}.
We compare our metric $e_\mathit{sgf}$ against the sum of absolute differences $e_\mathit{sad}=\sum\abs{e_\mathit{photo}}$, the absolute difference of gradient magnitude $e_\mathit{agm}=\abs{e_\mathit{gm}}$, the PatchMatch dissimilarity $e_\mathit{pm}$, and the original implementation. The dissimilarity in MeshStereo is calculated with Census-Transform. While OpenCV StereoBM uses $e_\mathit{sad}$ too, a different prefilter provided a better result for $e_\mathit{sad}$. All other metrics were evaluated without prefiltering. We omitted $\mathbf{e}_\mathit{gn}$ since the results were nearly indistinguishable from $e_\mathit{pm}$. The results are shown in \reftab{tab:MSB} and \reftab{tab:kitti}. As can be seen, our metric provides in all cases the best mean disparity error. \reffig{fig:Kitti} shows an example on the KITTI Stereo Benchmark. Please note for $e_\mathit{sgf}$, although the bicyclist is not well represented with MeshStereo, it is with StereoBM. Furthermore, in the background less incorrect (too close) disparities are calculated with our metric.

To support our second and third claim, we provide comparisons to a set of state-of-the-art VO and VIO approaches including DSO~\cite{DSO}, ORB-SLAM2~\cite{ORBSLAM2}, OKVIS~\cite{OKVIS} and SVO2~\cite{SVO2} on the EuRoC dataset. We implemented the different metrics in the optical flow frontend of Basalt and carried out a two-fold cross validation with hyperopt~\cite{hyperopt} to obtain suitable parameters for each metric. We use the Scharr-Operator~\cite{scharr} on the rotated patches to obtain the intensity gradients. 
We observed that using finite differences degraded the obtainable precision for this task. For disparity estimation finite differences are sufficient.

  In the case of DSO, we also show a modified version which replaces in the depth estimation the original patch similarity metric based on Brightness-Constancy-Assumption $e_\mathit{photo}$ with our $e_\mathit{sgf}$ term. \reffig{fig:DSO} shows an example for both on V1\_01 of the EuRoC dataset. For a fair comparison we disable the global bundle adjustment of ORB-SLAM2 and use Basalt purely in VIO mode. Furthermore, we evaluate the approaches, if provided, with the tailored parameters for the EuRoC dataset. 
  
  We report the mean ATE after alignment using \cite{zhang18alignment} for all the frames which have a pose estimate. We align DSO with a similarity transform and the stereo algorithms with a rigid transform.
  To achieve a more reliable error estimate we run the algorithms repeatedly for each scenario and average the results. We report also the number of successful trackings for each algorithm out of a total of 250. Tracking is considered failed if the maximum scale error is above \SI{1.5}{\meter} or the median scale error is greater than \SI{0.1}{\meter}. \reftab{tab:EurocATE} gathers the final results.

One can see that our modified DSO using the $e_\mathit{sgf}$ term for depth estimation performs better than the original DSO, having a lower average ATE. Furthermore, we observed an increase in successful tracking attempts by \SI{10}{\percent} on V1\_02 and V1\_03 which exhibit strong lighting changes and reduced variance in ATE.

Basalt achieves with all tested metrics excellent results. Presumably $e_\mathit{sgf}$ performs worse than our other derived metrics due to the more complex Jacobian, which is more difficult to optimize. Here, the simplifications of $e_\mathit{sgf2}$ and $e_\mathit{sgf3}$ payoff with $e_\mathit{sgf3}$ achieving the best result.

\bgroup
\newcommand{\qualitativeElemWidth}{8cm}
\renewcommand{\tabcolsep}{1pt}
\begin{figure}[h!]
	\centering
	\begin{tabular}{lc}
		\parbox[t]{3mm}{\rotatebox[origin=c]{90}{\small RGB}} & \includegraphics[width=\qualitativeElemWidth{},align=c]{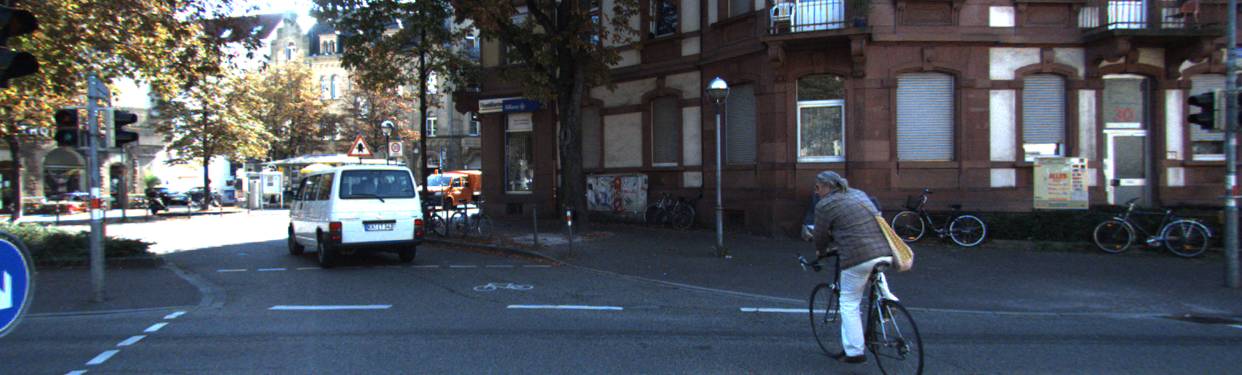} \\
\parbox[t]{3mm}{\rotatebox[origin=c]{90}{\small GT}} & \includegraphics[width=\qualitativeElemWidth{},align=c]{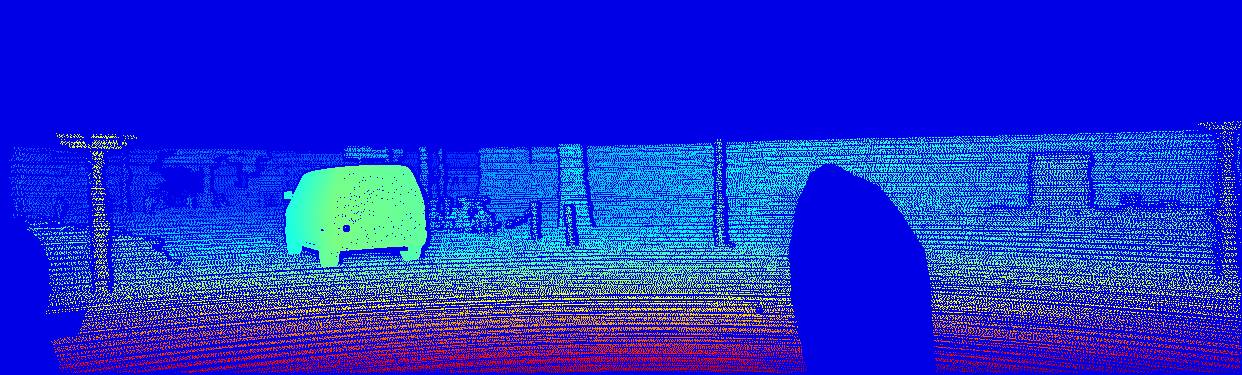}\\
		\parbox[t]{3mm}{\rotatebox[origin=c]{90}{\small Orig.}} & \includegraphics[width=\qualitativeElemWidth{},align=c]{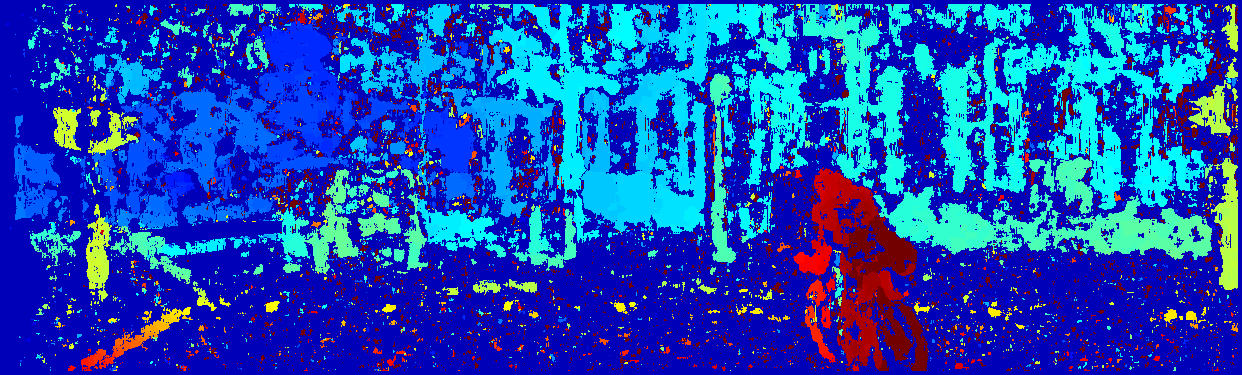} \\
		\parbox[t]{3mm}{\rotatebox[origin=c]{90}{\small $e_\mathit{pm}$}} &  \includegraphics[width=\qualitativeElemWidth{},align=c]{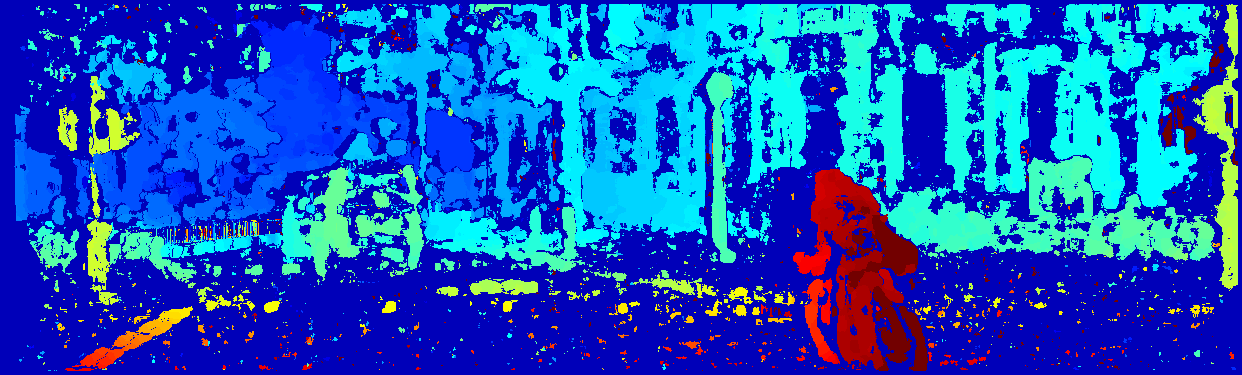} \\
		\parbox[t]{3mm}{\rotatebox[origin=c]{90}{\small $e_\mathit{sgf}$}} & \includegraphics[width=\qualitativeElemWidth{},align=c]{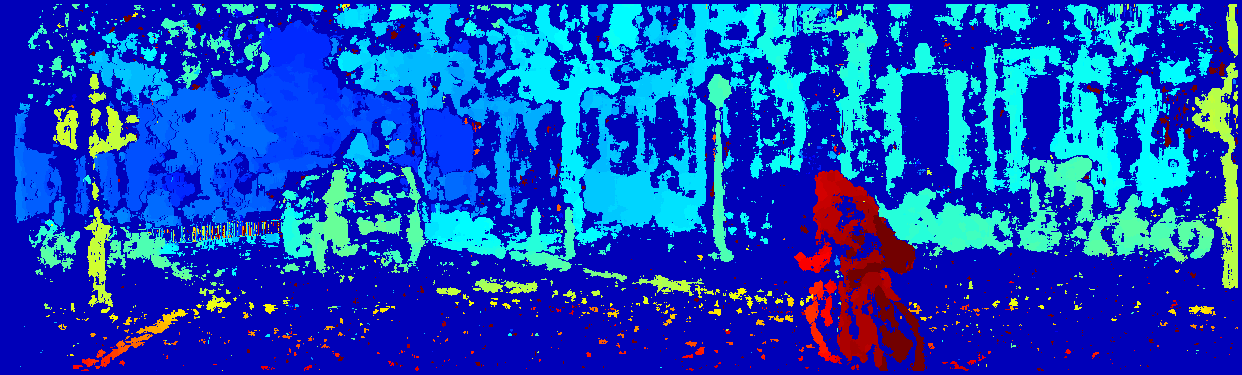} \\
		\parbox[t]{3mm}{\rotatebox[origin=c]{90}{\small Orig.}} & \includegraphics[width=\qualitativeElemWidth{},align=c]{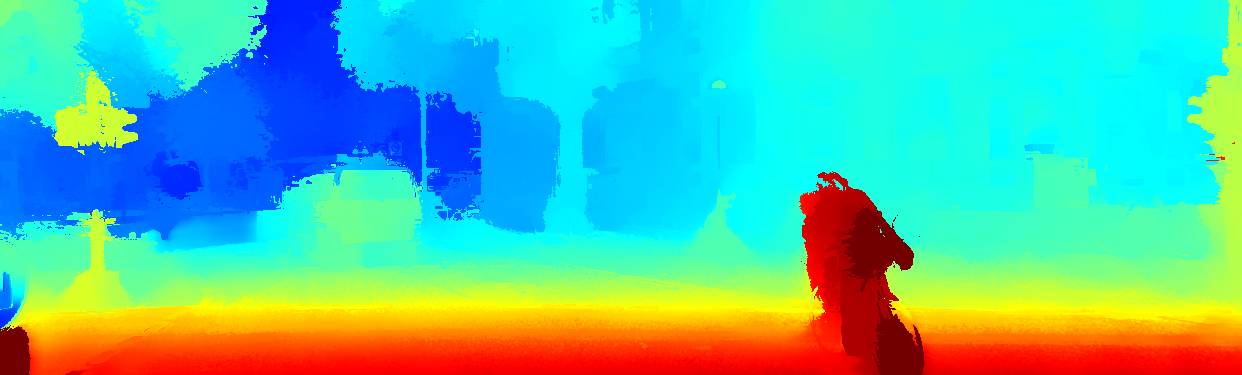}\\
		\parbox[t]{3mm}{\rotatebox[origin=c]{90}{\small $e_\mathit{pm}$}} & \includegraphics[width=\qualitativeElemWidth{},align=c]{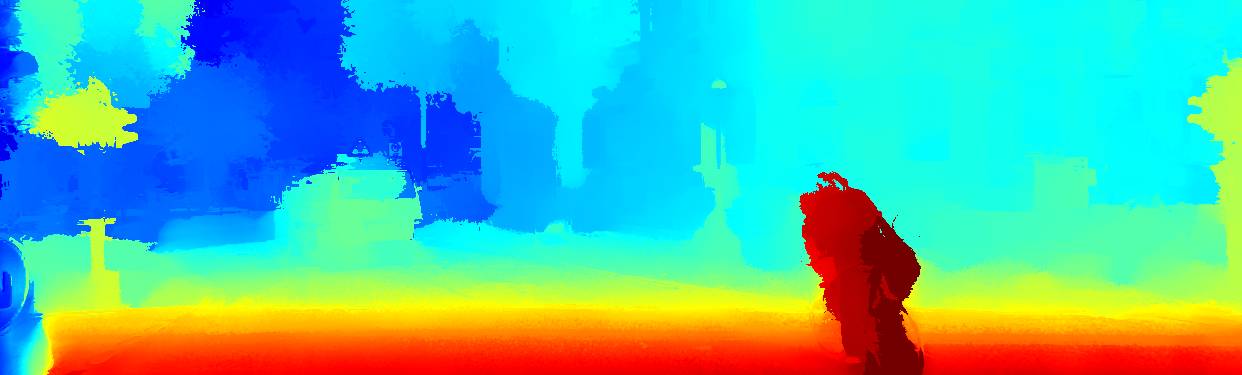}\\
		\parbox[t]{3mm}{\rotatebox[origin=c]{90}{\small $e_\mathit{sgf}$}} & \includegraphics[width=\qualitativeElemWidth{},align=c]{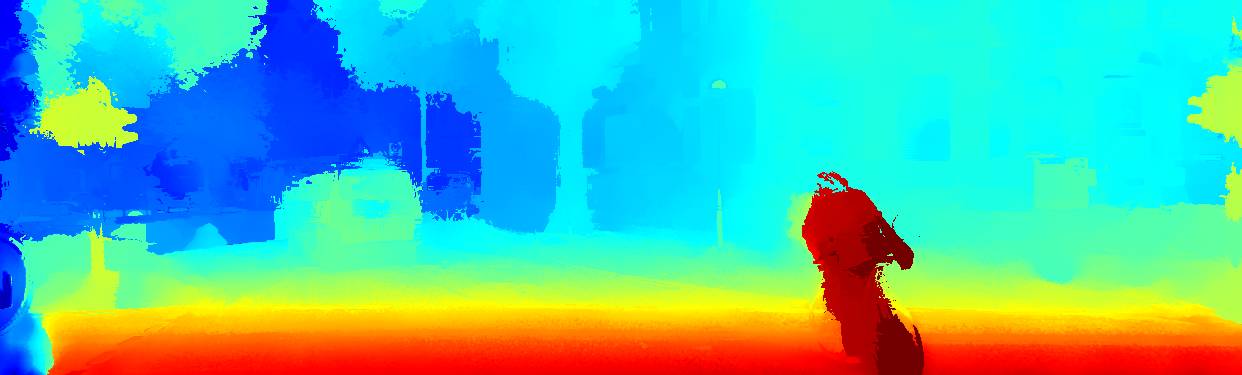}
	\end{tabular}
	\caption{Disparity comparison on image pair 2 of the KITTI Stereo 2015 Benchmark~\cite{kitti2015} for the original algorithms and the two best metrics.}
	\label{fig:Kitti}
\end{figure}
\egroup

\bgroup
\newcommand{\qualitativeElemWidth}{\linewidth} 
\renewcommand{\tabcolsep}{1pt}
\begin{figure}[h!]
	\centering
	\begin{tabular}{c}
		\includegraphics[trim=200 0 0 0,clip,width=\qualitativeElemWidth{},align=c]{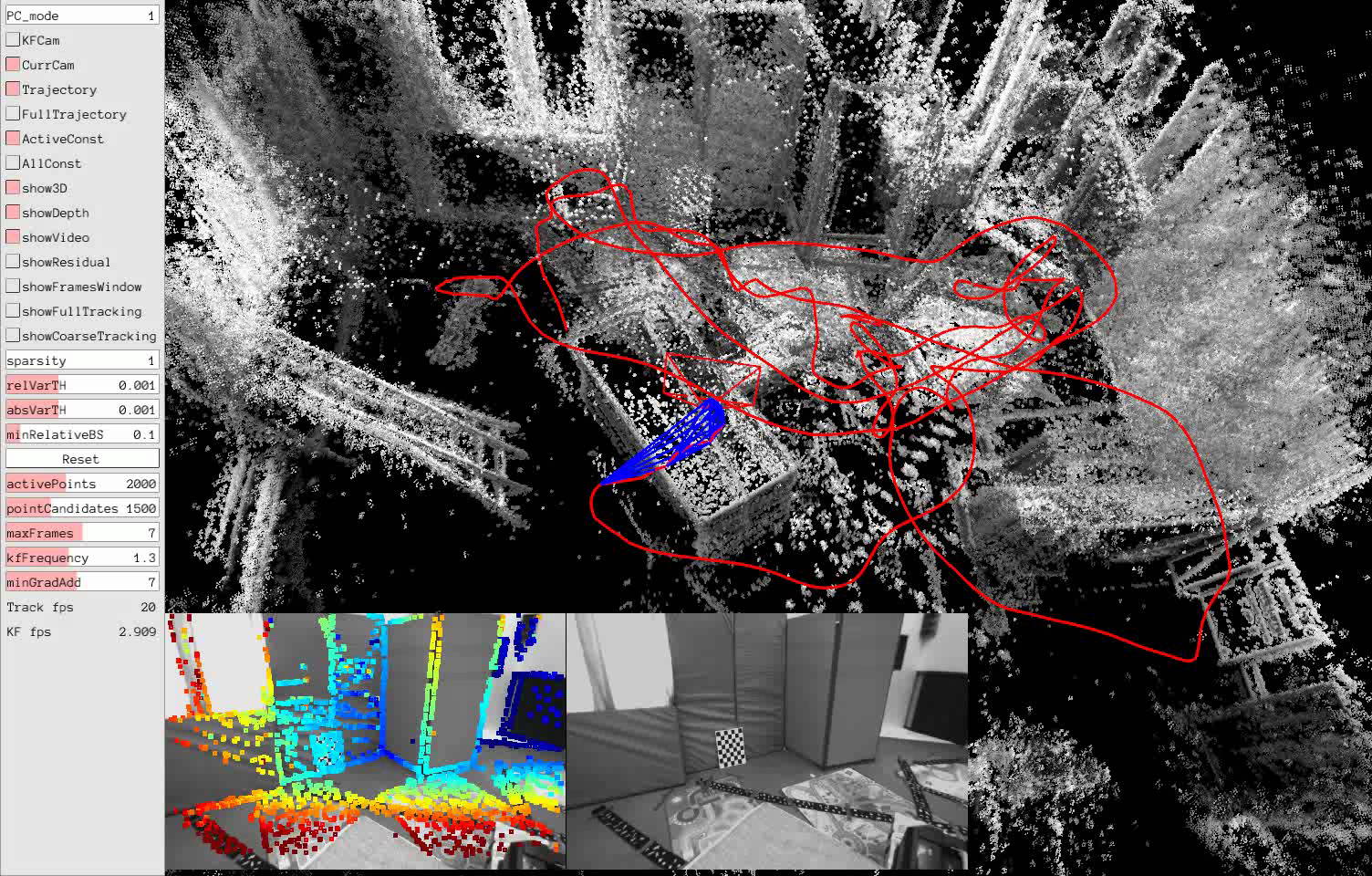} \\
		{\small Orig.} \\
		\includegraphics[trim=200 0 0 0,clip,width=\qualitativeElemWidth{},align=c]{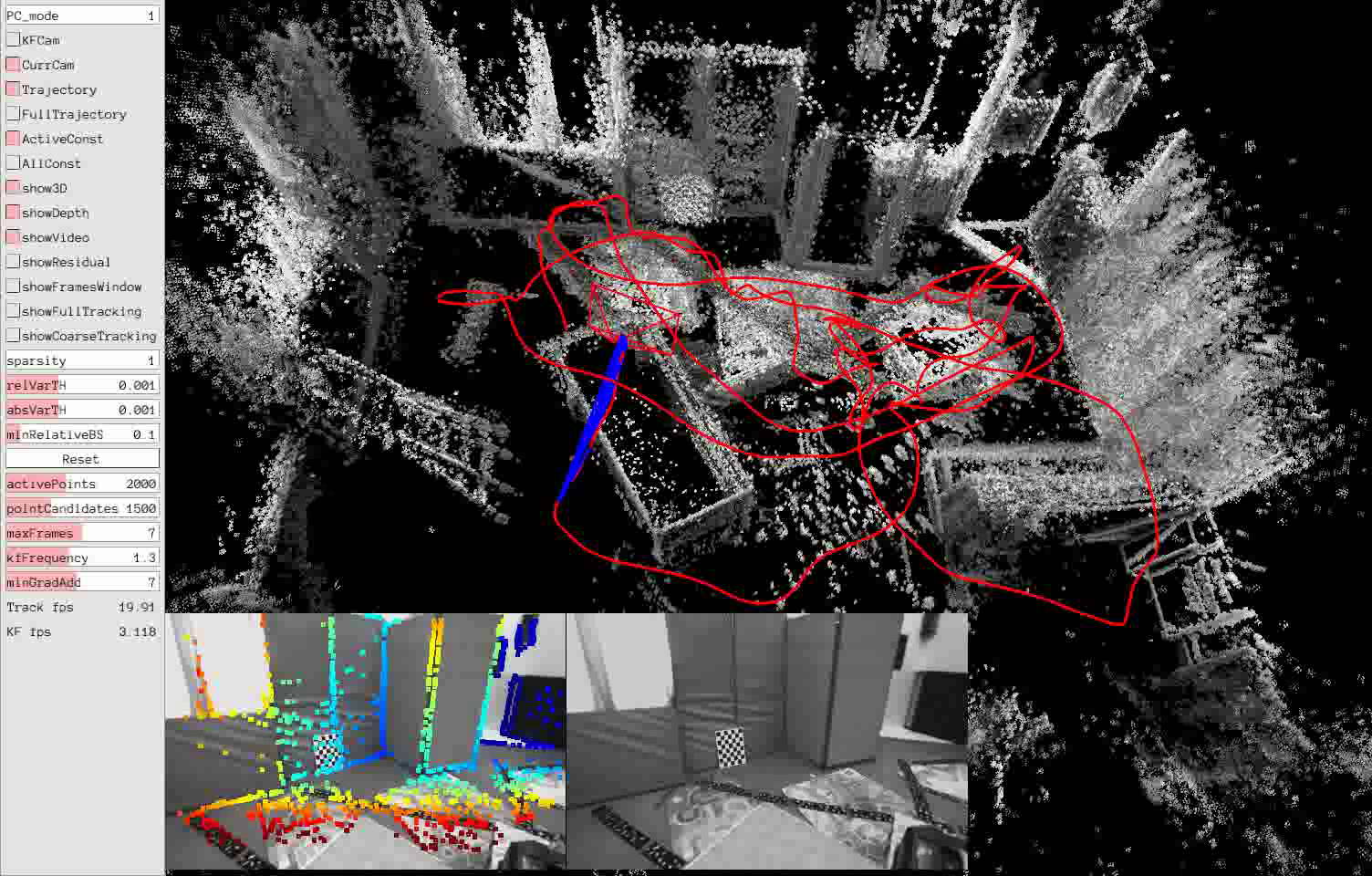}\\
		{\small $e_{sgf}$}
	\end{tabular}
	\caption{Resulting map and trajectory (red line) of DSO~\cite{DSO} w/o and with $e_{sgf}$ for depth estimation on V1\_01 of the EuRoC dataset~\cite{eurocMAV}. The reduced drift is clearly visible in the sharper edges and an reduction of double walls.}
	\label{fig:DSO}
\end{figure}
\egroup

\section{Conclusion}

In this paper, we proposed a new metric for direct image alignment that is useful for motion and stereo depth estimation. Our metric improves the gradient orientation metric proposed by Haber and Modersitzki~\cite{NGF} and integrates a magnitude-dependent scaling term. This improves the robustness of the image alignment and is beneficiary for stereo matching and visual odometry computation alike. We integrated and evaluated our approach in a multitude of settings showing that the proposed metric is better suited for disparity estimation than existing approaches and well suited for image alignment. Furthermore, our approach is easy to integrate into existing visual systems and thus can make a positive impact on various visual odometry, SLAM, or similar state estimation approaches.

\clearpage



\bibliographystyle{IEEEtranBST/IEEEtran}
\bibliography{bibliography}
\end{document}